\documentclass{article} 
\usepackage{lmrl2025_workshop,times}

\usepackage{amsmath,amsfonts,bm}

\def\eqref#1{equation~\ref{#1}}

\def\1{\bm{1}}

\DeclareMathAlphabet{\mathsfit}{\encodingdefault}{\sfdefault}{m}{sl}
\SetMathAlphabet{\mathsfit}{bold}{\encodingdefault}{\sfdefault}{bx}{n}

\usepackage{microtype}
\usepackage{graphicx}
\usepackage{booktabs} 
\usepackage{multirow}
\usepackage{adjustbox}
\usepackage{subcaption}
\usepackage{hyperref}
\usepackage{url}

\title{Tensor-DTI: Enhancing Biomolecular Interaction Prediction with Contrastive Embedding Learning}

\author{
\centerline{
\bf Manel Gil-Sorribes$^{1}$, \bf Júlia Vilalta-Mor$^{2,3}$, \bf Isaac Filella-Mercè$^{2}$, \bf Robert Soliva$^{5}$, }
\\
\centerline{
 \bf Álvaro Ciudad$^{1}$, \bf Víctor Guallar$^{2,4*}$, \bf Alexis Molina$^{1*}$
}
\\
\\
$^{1}$Nostrum Biodiscovery, 08029, Barcelona, Spain \\
$^{2}$Barcelona Supercomputing Center, 08034, Barcelona, Spain \\
$^{3}$Faculty of Pharmacy and Food Sciences, University of
Barcelona, 08028, Barcelona, Spain \\
$^{4}$Catalan Institution for Research and Advanced Studies (ICREA), 08010, Barcelona, Spain \\
$^{5}$Data Science Dpt., Almirall S.A., 08980, St. Feliu de Llobregat, Spain \\
*\texttt{victor.guallar@bsc.es, alexis.molina@nostrumbiodiscovery.com} 
}

\lmrlfinalcopy 
\begin{document}

\maketitle

\begin{abstract}
Accurate drug-target interaction (DTI) prediction is essential for computational drug discovery, yet existing models often rely on single-modality predefined molecular descriptors or sequence-based embeddings with limited representativeness. We propose Tensor-DTI, a contrastive learning framework that integrates multimodal embeddings from molecular graphs, protein language models, and binding-site predictions to improve interaction modeling. Tensor-DTI employs a siamese dual-encoder architecture, enabling it to capture both chemical and structural interaction features while distinguishing interacting from non-interacting pairs. Evaluations on multiple DTI benchmarks demonstrate that Tensor-DTI outperforms existing sequence-based and graph-based models. We also conduct large-scale inference experiments on CDK2 across billion-scale chemical libraries, where Tensor-DTI produces chemically plausible hit distributions even when CDK2 is withheld from training. In enrichment studies against Glide docking and Boltz-2 co-folder, Tensor-DTI remains competitive on CDK2 and improves the screening budget required to recover moderate fractions of high-affinity ligands on out-of-family targets under strict family-holdout splits. Additionally, we explore its applicability to protein-RNA and peptide-protein interactions. Our findings highlight the benefits of integrating multimodal information with contrastive objectives to enhance interaction-prediction accuracy and to provide more interpretable and reliability-aware models for virtual screening.
\end{abstract}

\section{Introduction}

The vast chemical space, estimated at up to \(10^{60}\) small molecules \citep{chemicalspace}, presents a major challenge for drug discovery, as practical exploration is constrained by synthesizability, stability, biological relevance, and the inherent difficulty of exploring such an immense space. Even if the exploration is limited to molecules satisfying Lipinski’s Rule of Five \citep{Lipinski}, the number of feasible drug-like molecules remains in the range of \(10^{12}\) to \(10^{23}\), making exhaustive screening infeasible. High-throughput experimental and virtual screening (HTS and HTVS) help navigate this space, but both remain limited by scalability constraints and predefined libraries. Experimental HTS is costly and typically limited to only tens to thousands of compounds (\(10^{1}\)-\(10^{3}\)), except for DEL-based HTS, which can explore much larger but structurally restricted linear libraries (\(10^{9}\)-\(10^{12}\)). \textit{In silico} HTVS methods based on molecular-modeling simulations, such as docking, generally scale only to a few million compounds (\(10^{6}\)). Meanwhile, the enlisted chemical space has grown exponentially, bolstered by combinatorial chemistry, with ultra-large libraries such as ENAMINE REAL \citep{enamine} containing over \(70\) billion readily synthesizable compounds and ZINC22 \citep{zinc22} offering access to more than \(97\) billion molecules. Thus, both HTS and HTVS, remain beyond the scope of exhaustively evaluating such vast chemical libraries.

Machine learning-based (ML-based) approaches have emerged as an alternative either by accelerating docking predictions (e.g., surrogate models such as \citep{surr} and \citep{sf}) or by bypassing the need for exhaustive evaluation (e.g., active learning models such as \citep{al}). Other ML-based models such as DiffDock \citep{diffdock} and TankBind \citep{tankbind} use geometric deep learning architectures, DiffDock via SE(3)-equivariant convolutional networks and TankBind via graph-based and trigonometry-aware networks, to predict binding poses with high efficiency. However, their dependence on large co-crystal datasets makes them vulnerable to data scarcity, limiting generalization to underrepresented protein families and unseen chemotypes. In contrast, molecule-protein interaction data is substantially more abundant with interaction datasets, like ChEMBL \citep{chembl}, PDBBind \citep{pdbbind}, and DUD-E \citep{dude}, providing a foundation for training predictive models, albeit with quality issues and information leakage between training and test sets, hindering model robustness \citep{leakproof}. This imbalance between scarce structural coverage and comparatively richer interaction data has motivated a shift toward sequence-based  drug-target interaction (DTI) models, which leverage sequence representations such as  protein language models like ESM \citep{esm} and SaProt \citep{saprot}, along with graph-based ligand encoders like GraphDTA \citep{graphdta} and HyperAttentionDTI \citep{hyperattention}, to enhance interaction prediction.

Despite these advances, current DTI models still face fundamental challenges in capturing the full complexity of biomolecular interactions. Many existing approaches rely solely on whole-protein embeddings, overlooking the importance of localized binding site information, which plays a critical role in molecular recognition and selectivity. While graph neural networks (GNNs) and transformer-based architectures have improved interaction modeling \citep{gnncti,transformercpi}, they often struggle to generalize to unseen drugs and targets due to their dependence on fixed molecular representations \citep{apdomain}. Contrastive learning frameworks such as ConPLex \citep{ConPlex} and PocketDTA \citep{pocketdta} have attempted to refine feature spaces by embedding proteins and drugs in a shared representation, but most fail to explicitly incorporate multimodal structural information, which is essential for capturing the nuances of binding affinities and selectivity. Additionally, generalization remains a major concern, as existing models often perform poorly on interactions beyond their training distribution, limiting their real-world applicability.

Motivated by the persistent challenges regarding missing binding site information, limited  generalization, and lack of multimodal integration in current DTI and drug-target affinity prediction (DTA) models, we introduce Tensor-DTI, a deep learning framework that integrates multimodal embeddings into a shared latent space and a siamese dual-encoder with contrastive learning to enhance DTI prediction. In addition, Tensor-DTI incorporates both global and localized structural features, leveraging pocket embeddings alongside protein and ligand representations to refine binding-site specificity when pocket information is available. By explicitly modeling binding pockets using PickPocket \citep{pickpocket}, a hybrid approach that integrates protein language models with structural message-passing networks \citep{gearnet}, Tensor-DTI provides a more interpretable and biologically grounded representation of molecular interactions. The model combines structural, chemical, and contextual information, enabling it to generalize across diverse biomolecular interactions, including peptide-protein-protein and RNA-protein interactions. This architecture allows Tensor-DTI to outperform existing sequence- and graph-based models on standard DTI and DTA benchmarks while offering insights into interaction specificity, making it a scalable and generalizable tool for drug discovery. In addition to benchmark evaluations, we assess the model’s capacity for hit recovery through a large-scale virtual screen of the Enamine REAL library on cyclin-dependent kinase 2 (CDK2) and through enrichment analyses on CDK2, acetylcholinesterase (AChE), and human monoamine oxidase A (MAO-A), where we compare Tensor-DTI rankings against Glide \citep{glide} (docking protocol in Section~\ref{glide_methods}) and Boltz-2 \citep{boltz2}.

\section{Results and Discussion}

We evaluate Tensor-DTI across multiple benchmarks, comparing it to competitive methods in both classification and affinity prediction tasks (DTI and DTA, respectively). Additionally, we assess its prospective applicability using recent leak-proof datasets. 

Ablation studies (see Appendix \ref{ablation}) identified pretrained molecular embeddings for drugs and structural embeddings for proteins as the best combination for DTI, while the optimal embeddings for DTA varied by dataset. A full description of all datasets, including preprocessing, splitting strategies, and dataset-specific details, is in Appendix \ref{databases}, while dataset sizes are in Appendix \ref{dataset_volumes}.

\subsection{Benchmarking Tensor-DTI}

To evaluate the predictive performance of Tensor-DTI in DTI scenarios, we conducted benchmarking experiments on multiple standard datasets. These included BIOSNAP, BindingDB, and DAVIS, alongside two additional BIOSNAP splits assessing generalization to unseen drugs and unseen targets. Notably, although all training splits were class-balanced, the test sets exhibited markedly different imbalance ratios. The BIOSNAP splits ($\sim$1:1), BindingDB ($\sim$1:6) and DAVIS ($\sim$1:19), reflecting a substantial predominance of negative pairs in these datasets (details in Table~\ref{tab:dataset_volumes}). The results, summarized in Table \ref{tab:model_performance}, provide a comparative assessment against established deep learning baselines and classical machine learning methods.

\begin{table}[h!]
\centering
\caption{Model Performance on standard DTI datasets. Each model for each dataset has been run 5 times. Performance is reported as the Area Under Precision Recall (AUPR) of the prediction. Metrics for models with $^{\dagger}$ are taken from ref. \citep{moltrans}. Ridge regression is not applicable to the Unseen Drugs dataset split because a distinct model is trained for each drug in the training set.}
\vspace{0.15in}
\label{tab:model_performance}
\begin{adjustbox}{max width=\textwidth}
\begin{tabular}{@{}lccccc@{}}
\toprule
\textbf{Model}        & \textbf{BIOSNAP}         & \textbf{BindingDB}      & \textbf{DAVIS}          & \textbf{Unseen Drugs}   & \textbf{Unseen Targets} \\ \midrule
Tensor-DTI            & $0.903 \pm 0.003$       & $0.699 \pm 0.002$       & $0.547 \pm 0.006$       & $0.888 \pm 0.002$       & $0.839 \pm 0.003$       \\
ConPLex               & $0.897 \pm 0.001$       & $0.628 \pm 0.012$       & $0.458 \pm 0.016$       & $0.874 \pm 0.002$       & $0.842 \pm 0.006$       \\
EnzPred-CPI           & $0.866 \pm 0.003$       & $0.602 \pm 0.006$       & $0.277 \pm 0.009$       & $0.844 \pm 0.005$       & $0.795 \pm 0.004$       \\
MolTrans              & $0.885 \pm 0.005$       & $0.598 \pm 0.013$       & $0.335 \pm 0.017$       & $0.863 \pm 0.005$       & $0.668 \pm 0.045$       \\
GNN-CPI$^{\dagger}$   & $0.890 \pm 0.004$       & $0.578 \pm 0.015$       & $0.269 \pm 0.020$       & $-$                     & $-$                     \\
DeepConv-DTI$^{\dagger}$ & $0.889 \pm 0.005$    & $0.611 \pm 0.015$       & $0.299 \pm 0.039$       & $0.847 \pm 0.009$       & $0.766 \pm 0.022$       \\
Ridge                 & $0.641 \pm 0.000$       & $0.516 \pm 0.000$       & $0.320 \pm 0.000$       & $N/A$                   & $0.617 \pm 0.000$       \\ \bottomrule
\end{tabular}
\end{adjustbox}
\end{table}

Tensor-DTI achieved the highest predictive performance across all datasets, with mean AUPR scores of  $0.903 \pm 0.003$ on BIOSNAP,  $0.699 \pm 0.002$ on BindingDB, and  $0.547 \pm 0.006$ on DAVIS. Notably, BIOSNAP exhibits a relatively well-characterized interaction landscape, primarily comprising known, high-confidence drug-target interactions. The model’s superior performance on BIOSNAP suggests its ability to effectively capture high-level interaction patterns, likely facilitated by contrastive embedding learning, which optimizes the separation of interacting and non-interacting pairs.

The more challenging BindingDB dataset, which encompasses a broader range of experimentally validated interactions across diverse small-molecule chemotypes, results in lower predictive performance for all models. Tensor-DTI maintains a robust performance margin over alternative deep learning models such as ConPLex ($+7.1$), MolTrans \citep{moltrans} ($+10.1$), and EnzPred-CPI \citep{enzpred_cti} ($+9.7$). The lower performance on DAVIS, where binding interactions are limited to kinase inhibitors, highlights the inherent challenge of predicting selective interactions within structurally conserved protein families.

Among competing methods, ConPLex performs well on BIOSNAP ($0.897 \pm 0.001$) but exhibits a significant drop on BindingDB ($0.628 \pm 0.012$) and DAVIS ($0.458 \pm 0.016$), suggesting a sensitivity to data heterogeneity and potential limitations in generalization beyond the training distribution. EnzPred-CPI and MolTrans show comparatively lower performance, particularly on DAVIS ($0.277$ and $0.335$, respectively), where kinase inhibitors exhibit complex binding profiles that are difficult to capture with purely sequence-based representations. Ridge regression, as expected, exhibits the lowest performance across all datasets, reinforcing the necessity of deep-learning-based representations for capturing the non-linear and high-dimensional features governing biomolecular interactions.

Beyond in-distribution benchmarking, we assessed Tensor-DTI’s capacity to generalize to novel drug-like molecules and previously unobserved protein targets. The unseen drug split evaluates the model’s ability to infer interactions for chemical entities that do not appear in the training set, whereas the unseen target split assesses generalization to proteins with no direct training exposure.

Tensor-DTI exhibits superior performance in the unseen drug scenario, with an AUPR score of $0.888 \pm 0.002$, and achieves $0.839 \pm 0.003$ in the unseen target scenario, showing comparable performance to ConPLex ($0.842 \pm 0.006$) as the difference lies within the margin of error. These results indicate that Tensor-DTI captures meaningful chemical and protein features, enabling it to extend beyond memorized interactions. ConPLex follows with $0.874 \pm 0.002$ for unseen drugs, further highlighting its competitive performance in generalization tasks.

MolTrans and DeepConv-DTI \citep{deepconv} show greater variability, particularly in the unseen target setting ($0.668 \pm 0.045$ and $0.766 \pm 0.022$, respectively), suggesting higher sensitivity to dataset distribution shifts. The relatively lower performance of MolTrans across both unseen drug and target scenarios underscores the challenge of extrapolating to unseen chemical scaffolds or protein families.

\subsection{Effectiveness of the Contrastive Learning Approach and Evaluation on DUD-E Dataset}\label{dude_contr}

To further assess the effectiveness of the contrastive learning approach employed in Tensor-DTI, we evaluated the model on the DUD-E dataset \citep{dude}, focusing on the kinase family. DUD-E provides property-matched decoys for each active compound. This creates a challenging test of whether the model captures true interaction signals beyond basic molecular similarity.

\begin{figure}[h!]
    \centering
    \includegraphics[width=\linewidth]{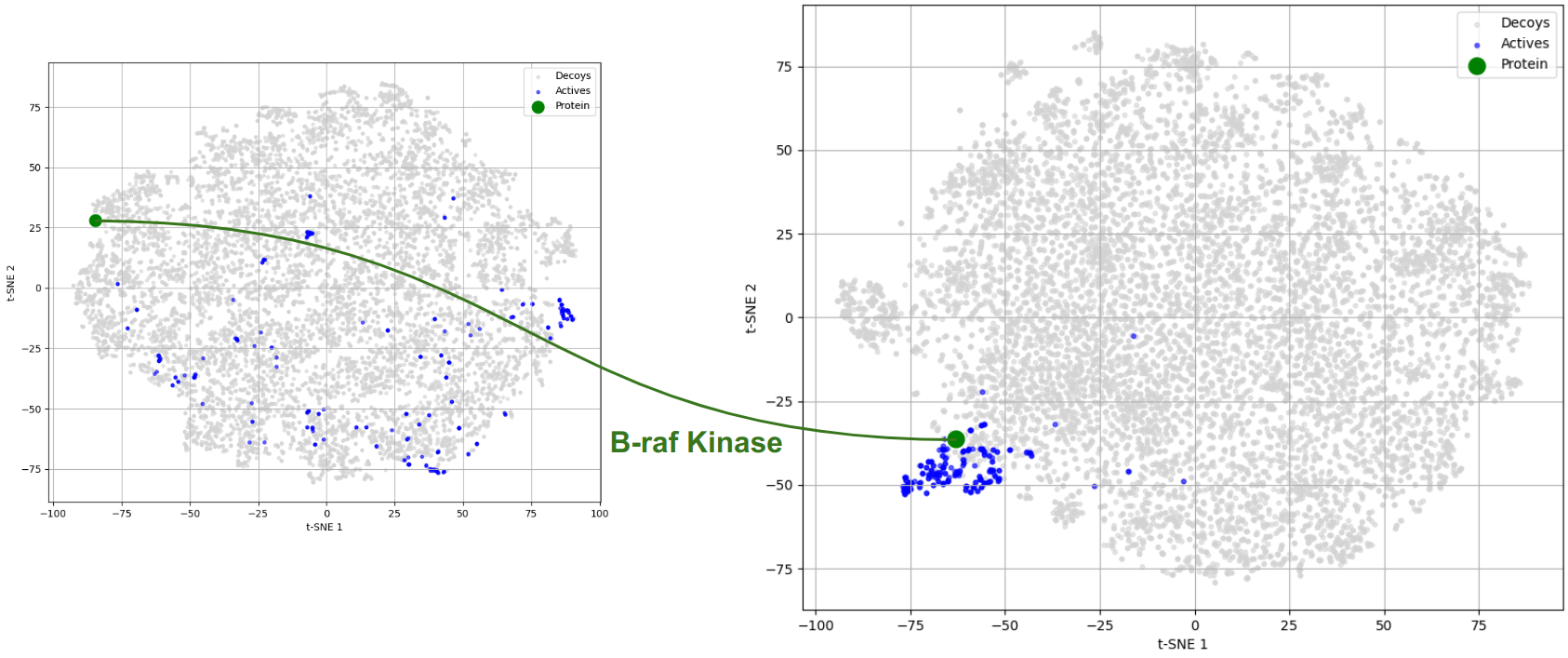}
    \caption{t-SNE visualization of protein and drug embeddings before (left plot) and after (right plot) applying Tensor-DTI with contrastive learning. The visualization corresponds to the B-raf Kinase protein, one of the targets from the test split.}
    \label{fig:dude_tsne}
\end{figure}

The performance of Tensor-DTI on this task is illustrated through a t-SNE visualization of the learned embeddings, with an example for one of the test proteins shown in Figure~\ref{fig:dude_tsne}. Prior to contrastive training, the embeddings of proteins and drugs lack clear separation between actives and decoys. After contrastive training, the model successfully clusters active drugs closer to their corresponding protein targets in the latent space, demonstrating improved discrimination between true binders and decoys. This structured embedding space suggests that the model effectively captures interaction-relevant molecular features. Tensor-DTI achieved an average AUPR of $0.686 \pm 0.006$, for all the test set, across five independent executions, confirming its strong capability in distinguishing actives from decoys.

These findings demonstrate the representational strength of Tensor-DTI, showing that contrastive learning not only improves predictive performance but also enhances interpretability. The structured latent space could be further leveraged for generative modeling, enabling the exploration of new active compounds based on proximity in the learned representation space.


\subsection{Tensor-DTI Performs Well on Affinity Prediction Datasets}\label{more_dta}

We tested Tensor-DTI on the Therapeutics Data Commons (TDC) DTI Domain Generalization \citep{tdc} (TDC-DG) benchmark, a challenging dataset for DTA prediction. The benchmark includes IC50 values from interactions patented between 2013 and 2018 as training data, while test interactions are from patents filed in 2019 and 2021. This setup demands strong out-of-domain generalization, simulating real-world applications where models predict unseen interactions based on historical data. To ensure robust performance, we evaluated multiple molecular and protein representations and found that, for this DTA dataset, Morgan fingerprints (MFPS) for small molecules and ESM-2 embeddings for proteins achieved the strongest performance. (see Appendix \ref{ablation}).

Following the data split strategy outlined in \citep{ConPlex}, we trained and evaluated Tensor-DTI in the DTA setting, which achieved a Pearson Correlation Coefficient (PCC) of $0.580 \pm 0.004$, demonstrating that our model is competitive with several state-of-the-art methods (Table \ref{tab:tdc_dg_results}).

\begin{table}[h!]
\centering
\caption{Comparison of Tensor-DTI performance on the TDC-DG benchmark.}
\vspace{0.15in}
\label{tab:tdc_dg_results}
\begin{tabular}{lc}
\toprule
\textbf{Model} & \textbf{PCC} \\
\midrule
Tensor-DTI & $0.580 \pm 0.004$ \\
ConPLex & $0.538 \pm 0.008$ \\
MMD & $0.433 \pm 0.010$ \\
CORAL & $0.432 \pm 0.010$ \\
ERM & $0.427 \pm 0.012$ \\
MTL & $0.425 \pm 0.010$ \\
\bottomrule
\end{tabular}
\end{table}

\subsection{DTI and DTA Assessments on Low-Leakage Datasets}

In order to evaluate drug-target interaction and affinity prediction models under minimized data leakage, we assessed performance across two curated datasets: PLINDER \citep{plinder} and LP-PDBBind \citep{leakproof}. These datasets were designed to reduce structural redundancy and prevent information leakage between training and test sets, making them valuable for assessing the generalization capacity of modern predictive models. Detailed performance comparisons across these datasets are provided in Appendix \ref{low_leak_results}.

For PLINDER, which contains only positive interaction pairs, we constructed negative examples and conducted two classification-based evaluations with different negative sampling strategies. In the first split, using only drug and protein embeddings, negative pairs were randomly selected from the same pool of drugs and proteins within each respective split, ensuring that non-interacting pairs were constructed exclusively from molecules present in that split. For the second split, we enforced structural dissimilarity between the original binding pockets and the pockets used for generating negative pairs.

The first approach resulted in an AUPR of \(0.785 \pm 0.002\), whereas the second achieved \(0.754 \pm 0.005\). This performance decrease when using highly dissimilar negative examples suggests that the model may leverage pocket similarity as a strong predictive heuristic. When this simplifying cue is removed by the experimental design, the model's ability to distinguish pairs is reduced. This highlights a potential limitation wherein DTI models may preferentially learn superficial correlations (e.g., pocket resemblance) over more complex molecular interaction features.

The further performance drop to $0.739$ AUPR when ablating the pocket embeddings in the dissimilar-negative setting reinforces this interpretation. It shows that, once pocket similarity can no longer be exploited, the model depends more heavily on explicit pocket features to resolve these more challenging classification cases. When both the heuristic and the explicit pocket information are removed, performance suffers substantially, indicating that pocket cues play a central role in the model’s decision process.

For LP-PDBBind, which is a DTA scenario, we tested the model respecting the splits proposed by the authors in \citep{leakproof}. Our model, which was trained to predict the \(K_d\), resulted in a PCC of \(0.565 \pm 0.004\) with a RMSE of \(1.620 \pm 0.024\). The model achieved lower PCC (\(0.528 \pm 0.013\)) and higher RMSE (\(2.122 \pm 0.032\)) for \(\Delta G\) than for \(K_d\), indicating greater noise and complexity in free energy predictions. To further investigate ligand-specific effects, we used the PDBBind-Opt \citep{optipdbbind} dataset separately to peptides (where we used \citep{peptidebert} for embedding generation) and small molecules. For peptides, the PCC reached \(0.679 \pm 0.014\), with a corresponding RMSE of \(1.175 \pm 0.020\). Meanwhile, for molecule-protein interactions, Tensor-DTI achieved a PCC of \(0.750 \pm 0.005\) with a RMSE of \(1.335 \pm 0.011\) on a random PDBBind-Opt split. However, when evaluated on \citet{leakproof}, which ensures no structural or sequence leakage between training and test sets, performance for molecule-protein interactions decreased to a PCC of \(0.493 \pm 0.005\) with a RMSE of \(1.545 \pm 0.006\). A summary of all benchmark results, including LP-PDBBind and PDBBind-Opt, is provided in Appendix~\ref{low_leak_results}.

\subsection{Tensor-DTI Allows Pocket Specificity with the Addition of Pocket Embeddings}

\begin{figure*}[h!]
    \centering
    \begin{subfigure}[t]{0.48\textwidth}
        \centering
        \includegraphics[width=\textwidth]{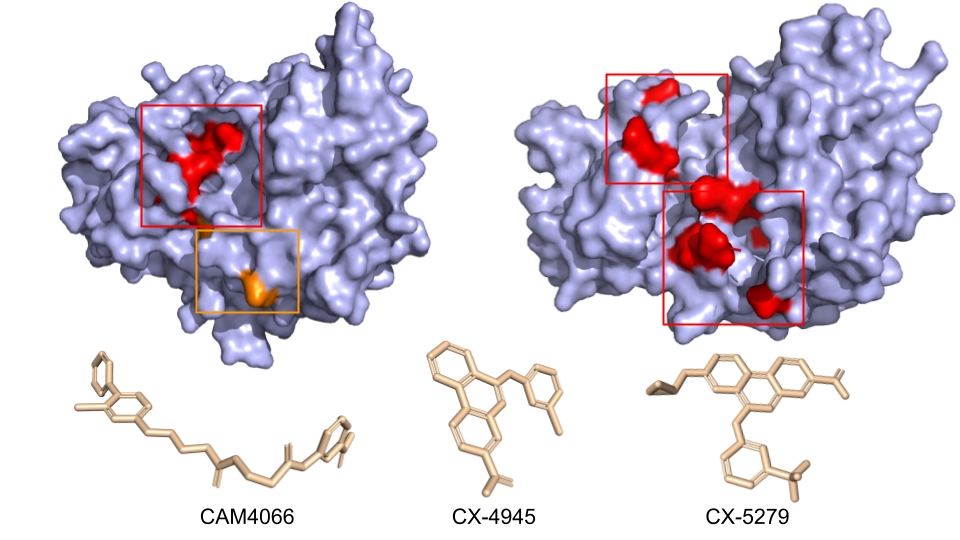}
        \caption{\textit{Left.} CDK2 with ATP binding site (red) and closed cryptic site (orange). \textit{Right.} Open cryptic cavity merging with ATP binding site (red). Structural configuration of the three binding compounds to the cryptic site. }
        \label{fig:cdk2}
    \end{subfigure}
    \hspace{0.02\textwidth}
    \begin{subfigure}[t]{0.48\textwidth}
        \centering
        \includegraphics[width=\textwidth]{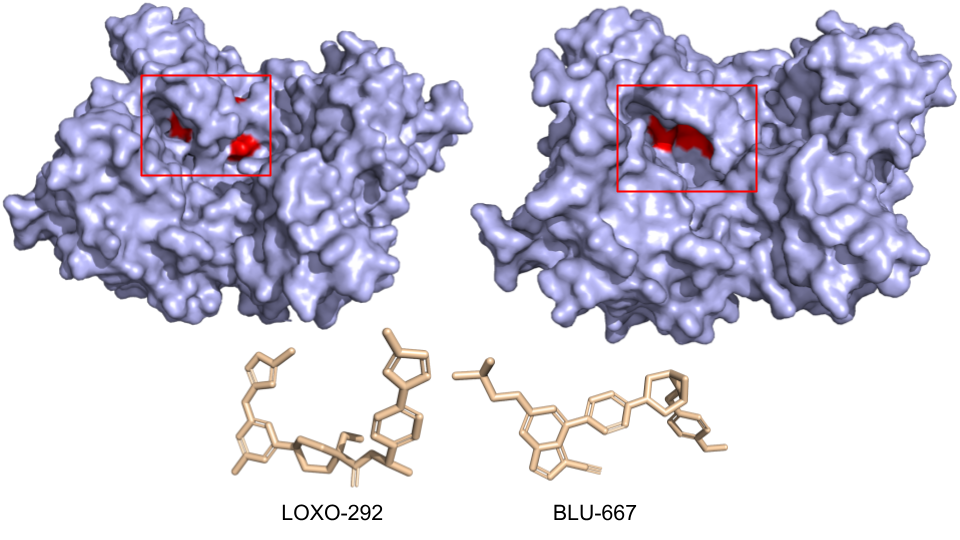}
        \caption{\textit{Left.} RET with cryptic binding site in apo state (red). \textit{Right.} Open cryptic cavity (red). Structural configuration of the two binding compounds to the cryptic site. }
        \label{fig:ret}
    \end{subfigure}
    \caption{Structural arrangements from CDK2 and RET kinases in holo and apo states for their corresponding cryptic pockets.}
    \label{fig:cdk2ret}
\end{figure*}

ML-based drug discovery often targets entire proteins, but modulating specific binding sites, including allosteric and cryptic pockets, offers greater therapeutic potential. Tensor-DTI enhances site-specific predictions by integrating pocket embeddings derived from PickPocket \citep{pickpocket}. PickPocket is trained on binding site data, which refines protein binding site representations using ESM-2 embeddings and GearNet-based structural message passing. These embeddings are combined with full-protein representations, improving the model’s ability to differentiate functionally relevant binding interactions from nonspecific contacts. The following experiments were performed using Tensor-DTI trained on the Plinder dataset.

\subsubsection{Assessment of Binding Across Cryptic Site Inhibitor Binding Predictions in CDK2 and RET Kinases}

CDK2 is a key regulator of the G1-to-S phase transition and is frequently hyperactivated in cancers such as breast, ovarian, and certain leukemias \citep{cdk2therapies}. Traditional ATP-competitive inhibitors struggle with selectivity and resistance mechanisms, making alternative binding site targeting an attractive approach. Thus, cryptic binding sites (CBSs) offer a promising avenue for kinase inhibitor development (Figure \ref{fig:cdk2}). To evaluate Tensor-DTI’s ability to distinguish cryptic sites from canonical ATP-binding pockets, we assessed ATP and selected CBS inhibitors across multiple conformational states of CDK2.

The model correctly rejected ATP binding to the closed ATP site in 3FWQ, reinforcing its ability to recognize steric constraints. In the cryptic conformation of 5CU3, where the CBS ligand CAM4066 is bound, the model successfully predicted CAM4066 as a binder in the cryptic pocket. This supports the model’s capability to recognize alternative binding pockets and ligand specificity. These results indicate the model’s sensitivity to structural context in cryptic binding scenarios.

Rearranged during transfection (RET) kinase is a critical target in thyroid and lung adenocarcinoma \citep{rettherapies}, but despite the approval of multi-tyrosine kinase inhibitors (MKIs) such as LOXO-292 (selpercatinib) and BLU-667 (pralsetinib), their long-term efficacy is often hindered by secondary mutations, off-target toxicity, and acquired resistance. To address these challenges, researchers have identified a cryptic binding site near the active site as a promising target for next-generation inhibitors (Figure \ref{fig:ret}). We evaluated Tensor-DTI’s ability to differentiate between the active site and CBS by predicting the binding of AMP, LOXO-292, and BLU-667 across multiple RET conformations.

In the case of RET, the cryptic and active sites are spatially close and share many of the same residues, making it difficult to distinguish between them. The model correctly predicted that LOXO-292 and BLU-667 bind to the cryptic site in the open conformation (7JU5), and it did so with higher confidence (See Section \ref{confidence_model}) than for the active site. However, it incorrectly predicted binding in the active site (2IVS), reflecting challenges in differentiating between highly similar pockets. Despite this, the model showed a clear preference for the cryptic site, suggesting it has learned to recognize features specific to cryptic accessibility.

AMP, a known binder to the RET active site (2IVS), was not correctly identified as such. Although the model failed to predict binding in 2IVS, it correctly rejected binding in the cryptic conformation (7JU5), with higher confidence in this non-binding prediction. This pattern highlights a consistent bias toward cryptic site recognition, potentially at the expense of accurately modeling ATP-competitive interactions. Overall, these results suggest that Tensor-DTI is better tuned to detect cryptic site features than subtle variations within canonical binding pockets, and that further refinement is needed to balance performance across both binding modes \citep{cryptic_site}.

\subsection{Generalization and Reliability in a Large-Scale CDK2 Virtual Screening Case Study}\label{cdk2_screening}

\begin{figure*}[t!p] 
\centering
\renewcommand\thesubfigure{\Alph{subfigure}} 

    \begin{subfigure}[b]{0.48\textwidth}
        \centering
        \includegraphics[width=\textwidth]{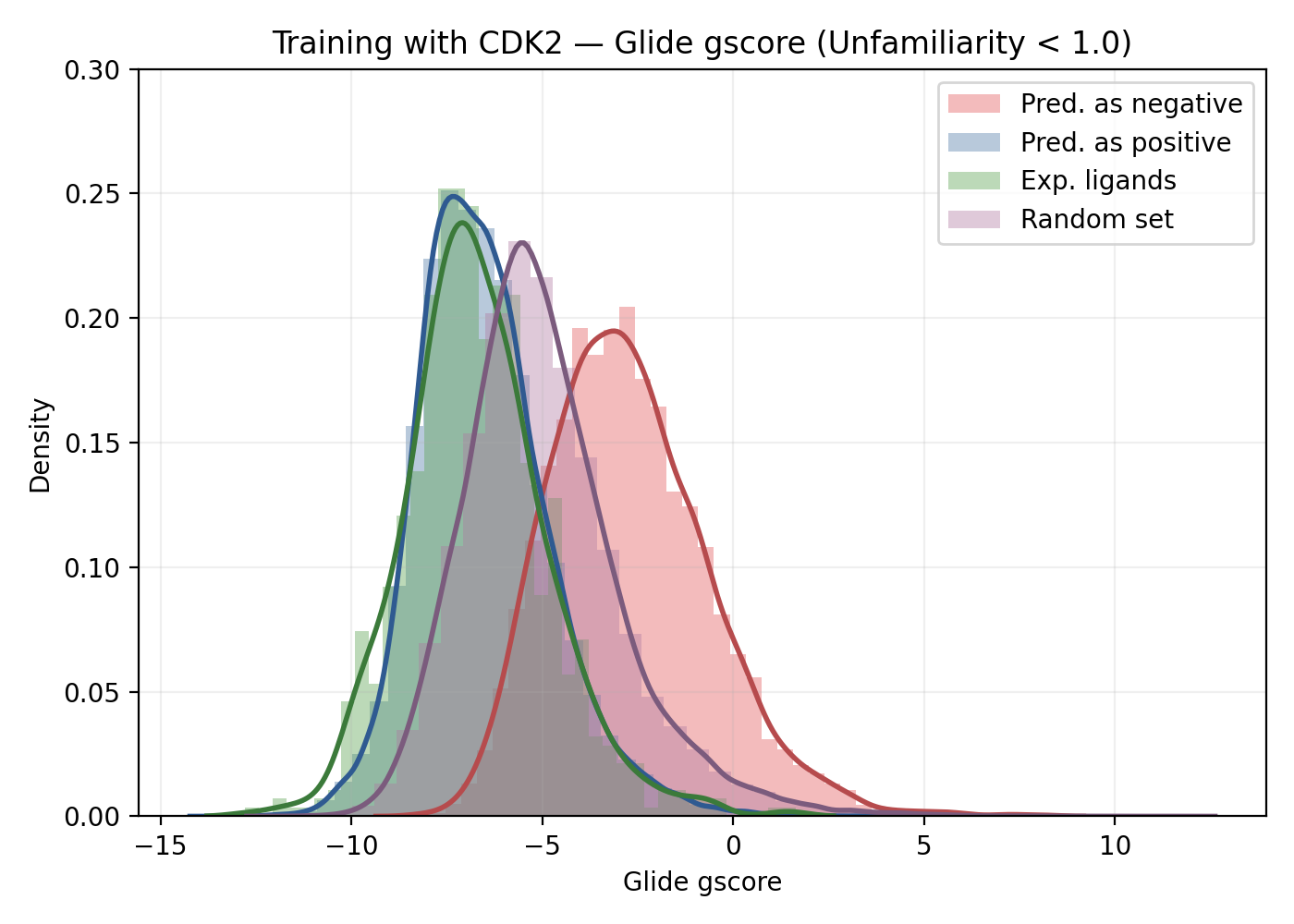}
        \caption{} 
        \label{fig:gscore_in}
    \end{subfigure}
    \begin{subfigure}[b]{0.48\textwidth}
        \centering
        \includegraphics[width=\textwidth]{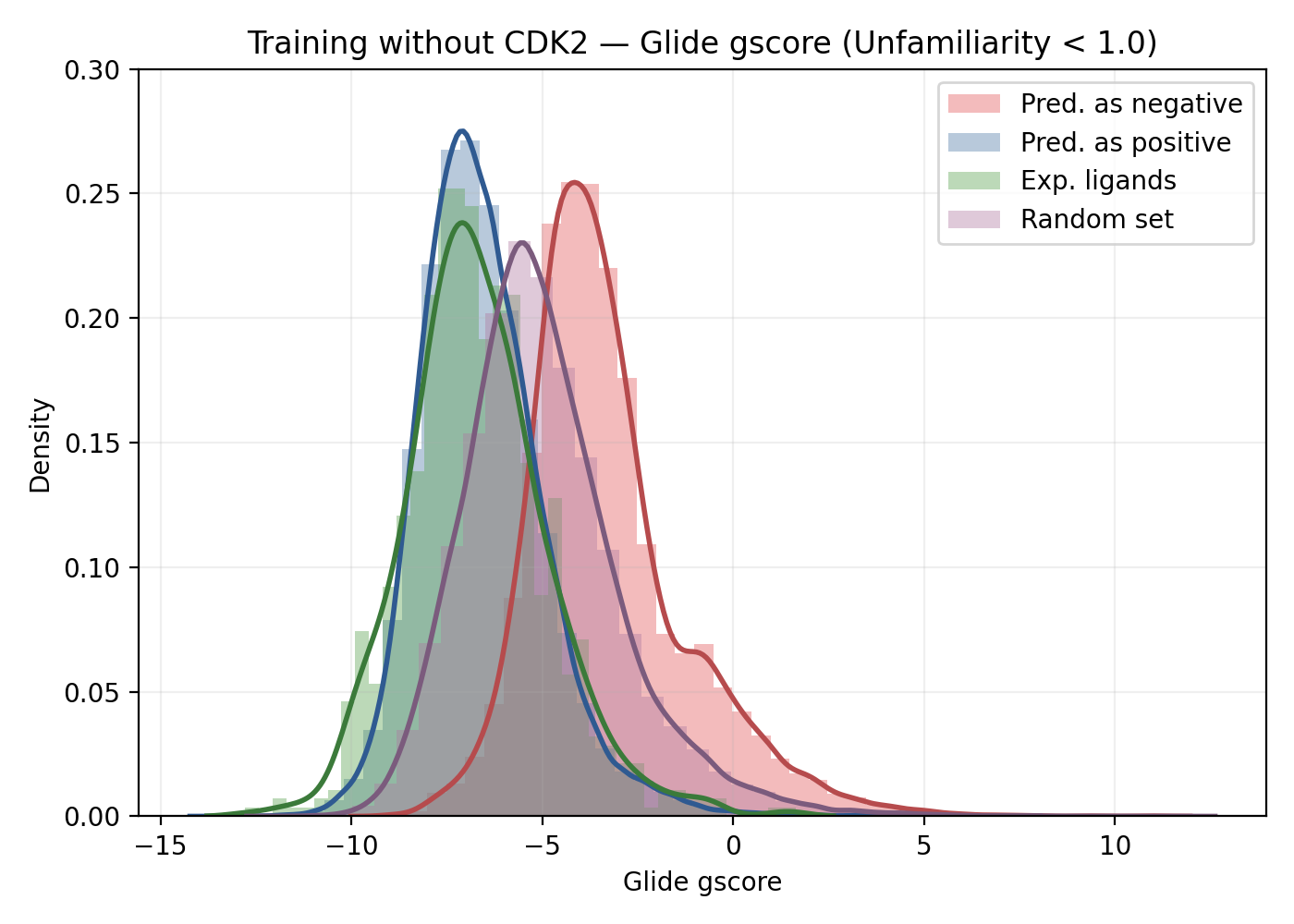}
        \caption{} 
        \label{fig:gscore_out}
    \end{subfigure}

    \par\smallskip 



    \begin{subfigure}[b]{0.48\textwidth}
        \centering
        \includegraphics[width=\textwidth]{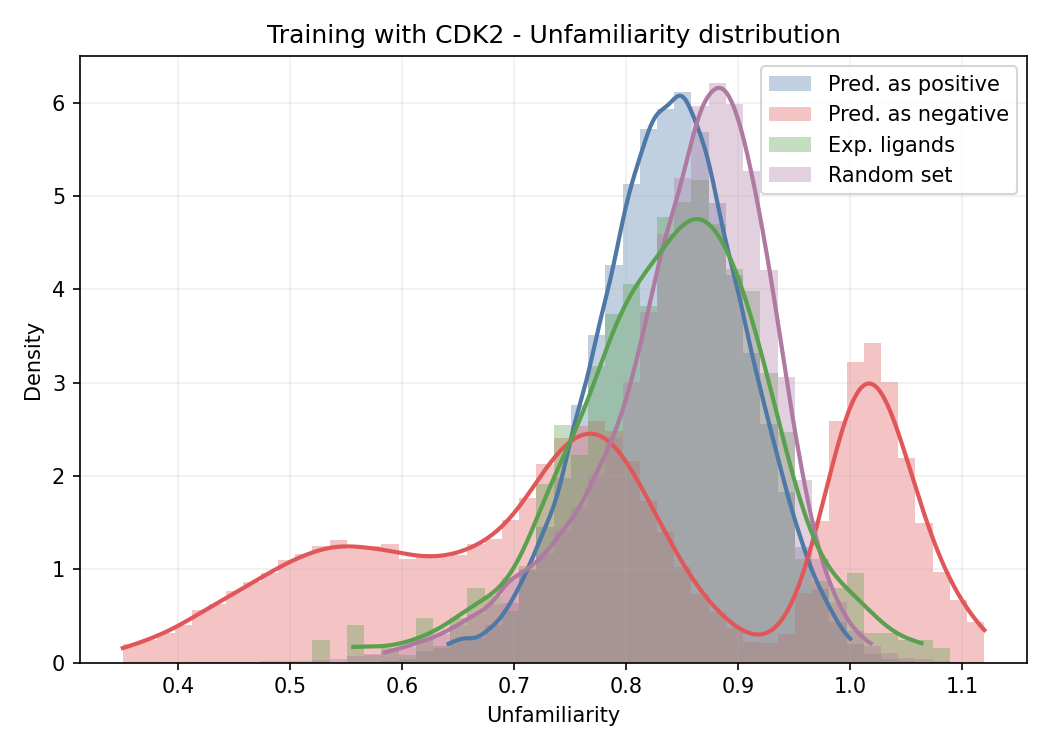}
        \caption{}
    \end{subfigure}
    \begin{subfigure}[b]{0.48\textwidth}
        \centering
        \includegraphics[width=\textwidth]{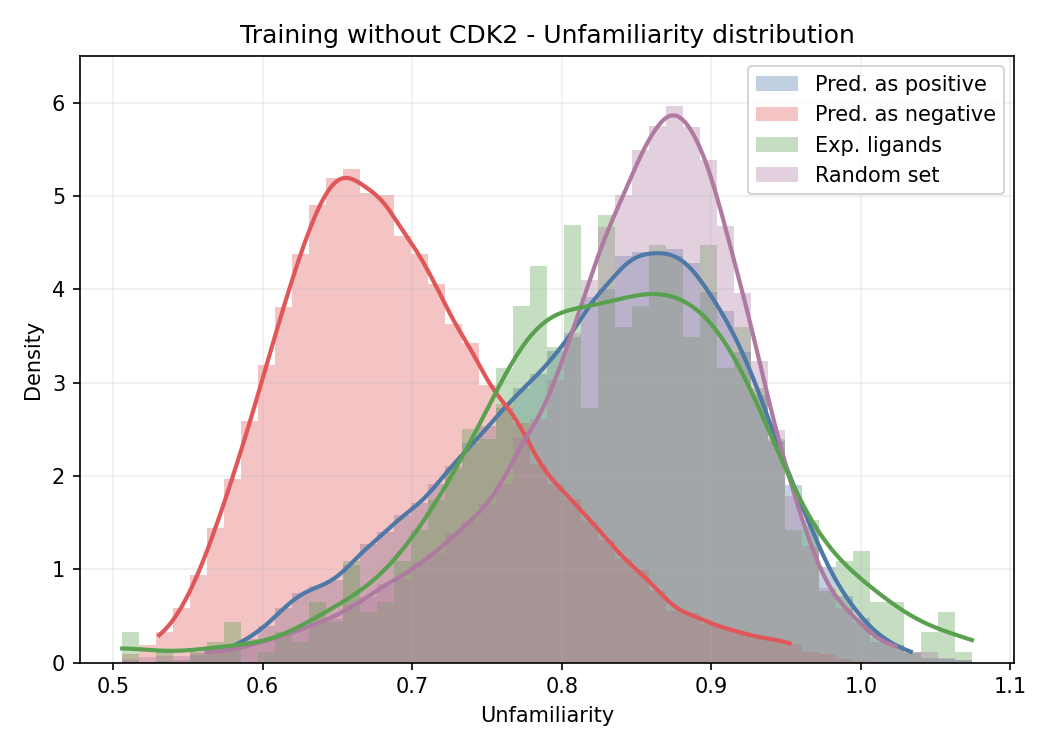}
        \caption{}
    \end{subfigure}

    \par\smallskip 

    \begin{subfigure}[b]{0.48\textwidth}
        \centering
        \includegraphics[width=\textwidth]{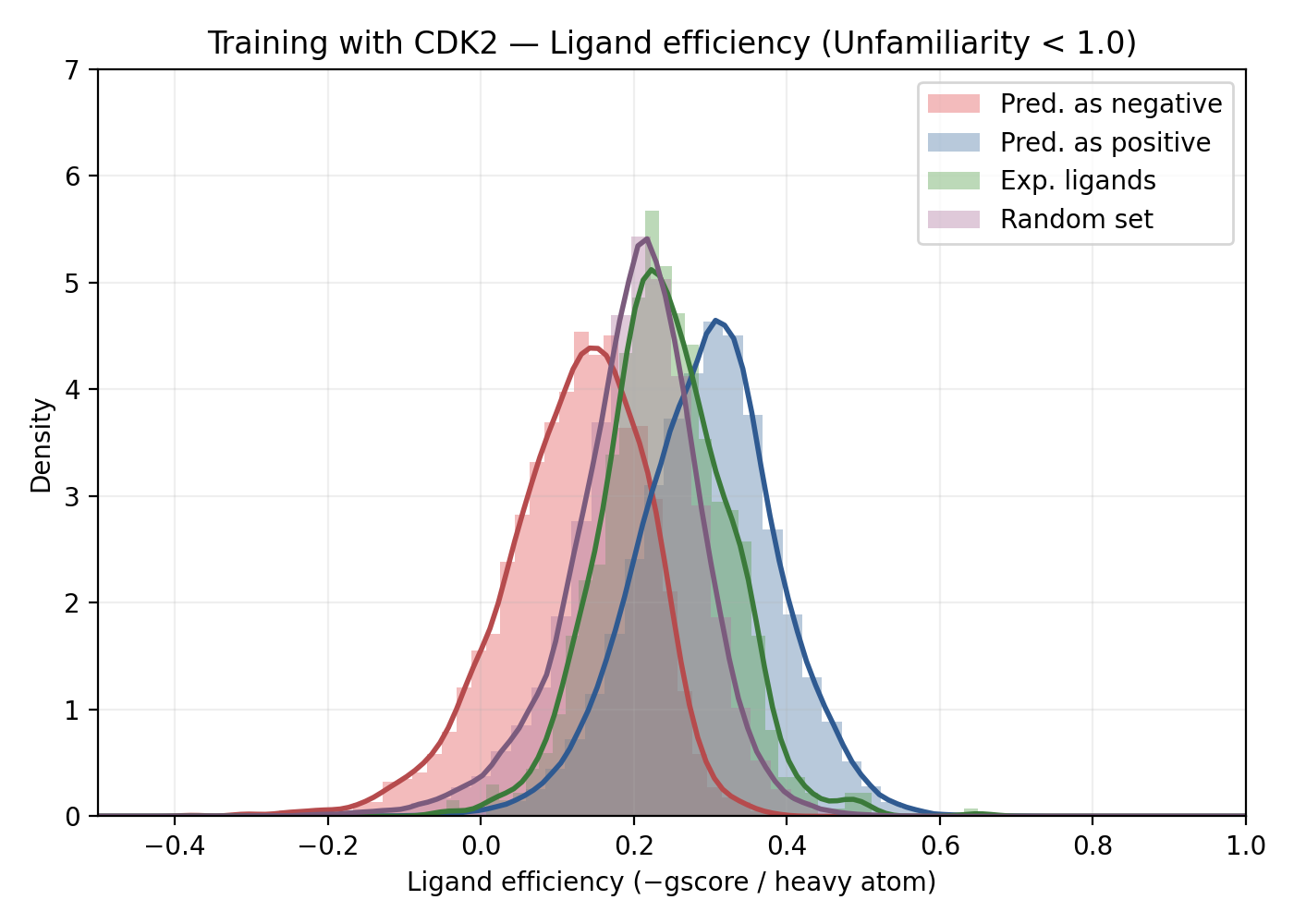}
        \caption{}
    \end{subfigure}
    \begin{subfigure}[b]{0.48\textwidth}
        \centering
        \includegraphics[width=\textwidth]{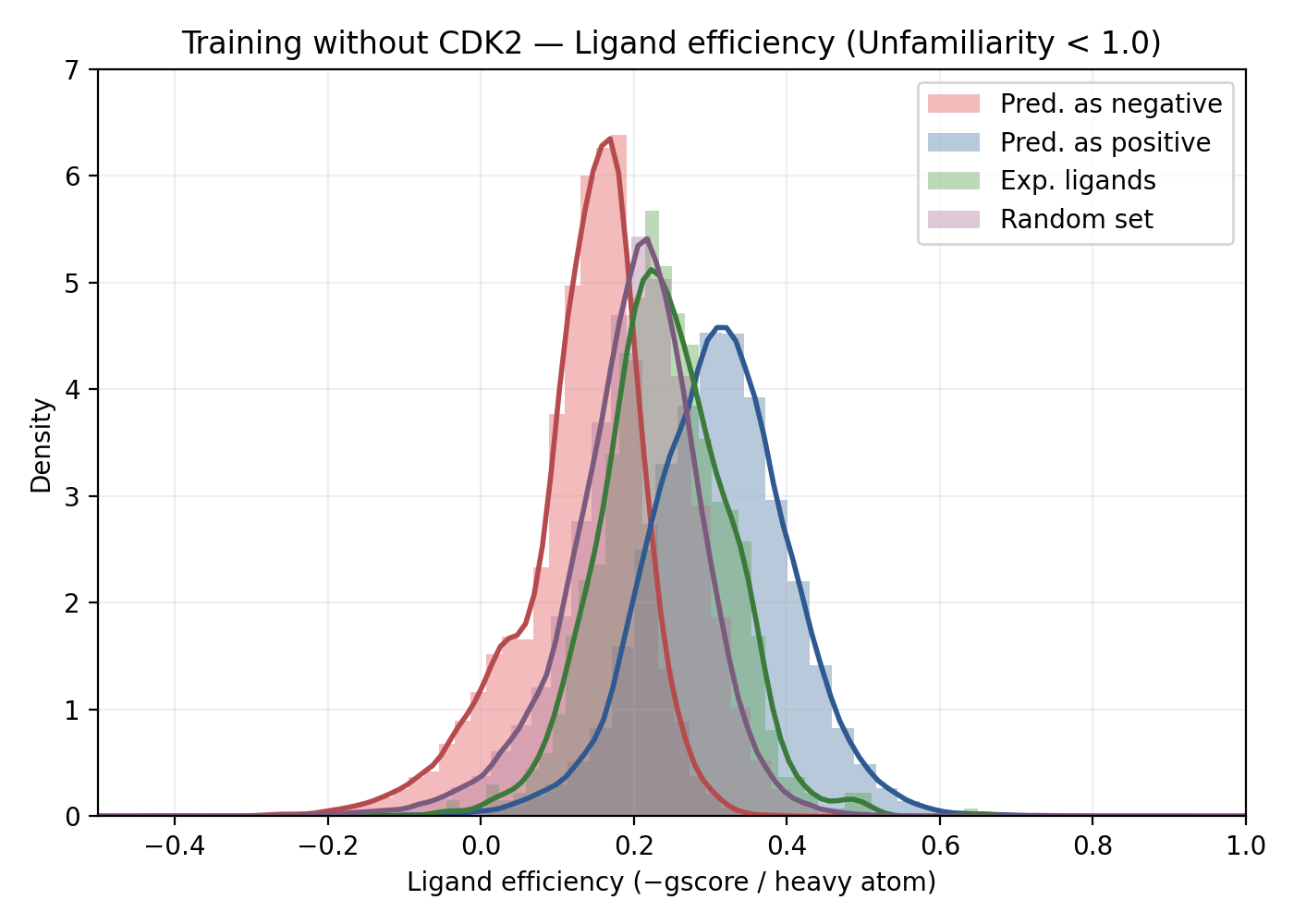}
        \caption{}
    \end{subfigure}
\caption{
CDK2 screening results with Tensor-DTI.
(A-B) Glide gscore distributions for models trained with and without CDK2, compared against experimental and random ligands.
(C-D) Unfamiliarity-based reliability distributions (filtered for unfamiliarity $<$ 1.0), showing that Tensor-DTI remains confident and chemically consistent even when CDK2 is excluded.
(E-F) Ligand efficiency distributions (-gscore per heavy atom), illustrating that the model preserves balanced, size-normalized scoring behavior.}
\label{fig:cdk2_results}
\end{figure*}

To evaluate Tensor-DTI's capacity for chemical generalization in realistic discovery settings, we conducted a large-scale virtual screen targeting the orthosteric site of CDK2. As a well-characterized kinase with well-defined structural features, CDK2 serves as an ideal benchmark for quantitatively comparing predicted interaction patterns against established experimental trends. We processed the Enamine REAL 5B library against the CDK2 target, generating embeddings and running inference using two model configurations: one trained with CDK2 data and one without. From the resulting predictions, we isolated the top 100\,000 highest-scoring molecules (putative actives) and the bottom 100\,000 lowest-scoring molecules (confident non-binders) to analyze the model's discriminatory power, as measured using Glide docking as the oracle. Figure~\ref{fig:cdk2_results} visualizes four distinct populations: (1) Predicted Positives, (2) Predicted Negatives, (3) known Experimental Ligands, and (4) a Random Set from the Enamine library.

To further assess reliability, we employed an unfamiliarity metric derived from our molecular autoencoder. Following the framework of \citet{dl_edge}, this metric quantifies the distance of a compound from the model’s learned chemical manifold, where high values indicate less trustworthy, out-of-distribution regions. For this analysis, we retained only compounds falling within the reliable region of the manifold (unfamiliarity $<$ 1.0; see Section~\ref{confidence_model}). We applied two filters to these sets: the availability of a pre-computed Glide gscore and an unfamiliarity score $<$ 1.0.

The initial populations consisted of 100\,000 predicted positives, 100\,000 predicted negatives, 85\,000 random compounds, and 817 experimental ligands. The filtering workflow and the resulting dataset sizes for each group are summarized in Table~\ref{tab:cdk2_filter_stats}.

\begin{table}[h!]
\centering
\caption{
Dataset sizes after applying the two reliability filters used in the CDK2 
screen: (i) availability of a valid Glide gscore and (ii) unfamiliarity 
$< 1.0$. Values correspond to the final populations analyzed throughout 
Figures~\ref{fig:cdk2_results}A-B and E-F. Figure~\ref{fig:cdk2_results} C-D include all the compounds.}
\label{tab:cdk2_filter_stats}

\begin{tabular}{lrr}
\toprule
\textbf{Population} & 
\textbf{Valid Gscore (docked)} & 
\textbf{Unf $<$ 1.0} \\
\midrule
\multicolumn{3}{l}{\textbf{Trained With CDK2}} \\
\midrule
Pred.\ Negatives      & 7\,313   & 5\,306 \\
Pred.\ Positives      & 76\,882  & 76\,518 \\
Experimental Ligands  & 817      & 782 \\
Random Set            & 85\,661  & 84\,722 \\

\midrule

\multicolumn{3}{l}{\textbf{Trained Without CDK2}} \\
\midrule
Pred.\ Negatives      & 9\,917   & 9\,908 \\
Pred.\ Positives      & 79\,125  & 78\,261 \\
Experimental Ligands  & 817      & 782 \\
Random Set            & 85\,661  & 84\,722 \\
\bottomrule
\end{tabular}
\end{table}

In both configurations, whether CDK2 was included in training or withheld, Tensor-DTI successfully recovered the expected activity landscape. When trained with CDK2, the predicted actives exhibited Glide gscores (protocol in \ref{glide_methods}) that overlapped with experimental ligands and showed a clear left-shift relative to random compounds (Figure~\ref{fig:cdk2_results}A) and predicted negatives. Notably, even when CDK2 was excluded from training (Figure~\ref{fig:cdk2_results}B), the activity landscape showed the same trend. This demonstrates robust transferability across related kinases.

We also examined ligand efficiency as a size-normalized measure of binding potential. In both training regimes (Figures~\ref{fig:cdk2_results}E-F), Tensor-DTI reproduced the general LE profile of kinase ligands. Predicted positives showed a consistent right-shift toward higher efficiencies compared to the experimental set, while predicted negatives and the random set centered lower. This suggests the model preferentially ranks compact, energetically favorable chemotypes.

Within this reliable regime, Tensor-DTI clearly distinguished actives, inactives, and random ligands. Figures~\ref{fig:cdk2_results}C-D show the full unfamiliarity 
distributions for all evaluated compounds, independent of any docking or unfamiliarity-threshold filtering. When CDK2 was included in training, predicted actives clustered around unfamiliarity values consistent with experimental compounds. Excluding CDK2 caused a slight shift toward higher unfamiliarity, yet the distribution retained its shape and separation. This behavior reflects the “edge of chemical space” phenomenon \citep{dl_edge}, where prediction quality gradually decays but remains interpretable up to a soft boundary.

We attempted a parallel screening campaign using the pocket-aware Tensor-DTI variant, however, convergence proved unstable. Inspection revealed that the available pocket-level dataset was insufficient to support generalization at inference scale. This was evidenced by broader, noisier Glide gscore distributions and systematically higher unfamiliarity values, indicating the model was operating outside its learned structural domain.

Overall, these experiments highlight Tensor-DTI’s ability to generalize across structurally related proteins. Even without direct training examples, the model identified relevant binders relevant binders. The unfamiliarity filter served as an effective quality control mechanism, ensuring the analysis reflected robust, in-domain behavior.

\subsection{A Comparative Enrichment Analysis of Tensor-DTI, Glide, and Boltz-2}\label{sec:enrichment}

We compared early retrieval enrichment for three ranking strategies on CDK2, AChE (UniProt: P21836), and MAO-A (UniProt: P21397): Glide gscore \citep{glide} (docking protocol described in Section~\ref{glide_methods}), Boltz-2, and Tensor-DTI. The former was trained on SMPBind I, with CDK2 variants, including (Tensor-DTI-c) or excluding (Tensor-DTI-nc) CDK2 interactions from training, and a single variant for AChE and MAO-A excluding all interactions related to the target and its protein family. For each target, true active hits are molecules with experimentally measured affinities, with higher affinity binders representing the most desirable hits. Before evaluation, we confirmed that no true active appeared in SMPBind I so that enrichment reflects genuine generalization rather than training overlap. For each method and target, we evaluated early enrichment using two metrics: $k$\% actives recovered (AR), the percentage of the ranked library that must be screened to recover fixed percentage of actives, and Top-k, the fraction of actives contained within the top portions of the ranked library. Both summaries are monotonic transforms of standard enrichment factors at fixed cutoffs. Full details of the ranking protocol, active set alignment, and the mapping to enrichment factor style metrics are given in Appendix~\ref{enrichment}.

\vspace{0.5em}

\begin{table}[h!]
\centering
\caption{CDK2 enrichment comparison. Each value shows the percentage of the ranked library required (ascending order) to recover $k$\% of all experimental actives, regardless of potency ranking. Lower values indicate better enrichment. Tensor-DTI results are shown for models trained with CDK2 (Tensor-DTI-c) and without CDK2 (Tensor-DTI-nc).}
\begin{adjustbox}{max width=\textwidth}
\begin{tabular}{lcccccc}
\toprule
\textbf{$k$\% AR} & \textbf{\# actives} & \textbf{Glide gscore} & \textbf{Boltz-2} & \textbf{Tensor-DTI-c} & \textbf{Tensor-DTI-nc} & \textbf{Random} \\
\midrule
1\%   & 8   & 0.57 & 0.45 & 1.63 & 2.00 & 1.00 \\
5\%   & 40  & 2.32 & 2.01 & 3.95 & 5.34 & 5.00 \\
20\%  & 160 & 11.94 & 8.04 & 10.93 & 14.15 & 20.00 \\
50\%  & 398 & 31.76 & 22.29 & 25.44 & 30.41 & 50.00 \\
100\% & 796 & 99.38 & 97.78 & 84.62 & 89.35 & 100.00 \\
\bottomrule
\end{tabular}
\end{adjustbox}
\label{tab:cdk2_enrichment}
\end{table}

\begin{table}[h!]
\centering
\caption{CDK2 enrichment comparison. Each value indicates the percentage of the ranked compound library that must be taken (in ascending order) to recover the corresponding fraction of experimentally validated binders. Lower values therefore denote earlier recovery and better enrichment. Tensor-DTI results are shown for models trained with CDK2 (Tensor-DTI-c) and without CDK2 (Tensor-DTI-nc).}
\begin{adjustbox}{max width=\textwidth}
\begin{tabular}{lcccccc}
\toprule
\textbf{Top-$k$} & \textbf{\# actives} & \textbf{Glide gscore} & \textbf{Boltz-2} & \textbf{Tensor-DTI-c} & \textbf{Tensor-DTI-nc} & \textbf{Random} \\
\midrule
1\%   & 8   & 80.92 & 28.41 & 46.55 & 51.11 & 88.90 \\
5\%   & 40  & 92.23 & 30.06 & 56.09 & 56.66 & 95.00 \\
20\%  & 160 & 99.26 & 82.81 & 79.24 & 89.35 & 96.20 \\
50\%  & 398 & 99.26 & 87.75 & 80.63 & 89.35 & 96.60 \\
100\% & 796 & 99.38 & 97.78 & 84.62 & 89.35 & 96.90 \\
\bottomrule
\end{tabular}
\end{adjustbox}
\label{tab:cdk2_enrichment}
\end{table}

On CDK2, Boltz-2 provides the strongest early enrichment when we measure the screening budget needed to recover a given fraction of binders. It reaches one, five, twenty, and fifty percent of the known actives after testing a smaller fraction of the library than any other method. Tensor-DTI c is consistently second best in this view and requires markedly fewer compounds than either docking or Boltz-2 to recover the full set of actives, which shows that its global ordering of ligands produces the shortest tail. In the complementary top-k view, Glide gscore attains the lowest recovery for a fixed top fraction of the library on CDK2, while Tensor-DTI c and Tensor-DTI nc remain competitive and clearly outperform random ranking.

Comparing Tensor-DTI-c and Tensor-DTI-nc on CDK2 quantifies unseen target generalization. Removing CDK2 from training weakens early enrichment, especially at the very first percent of recovered actives, yet Tensor-DTI-nc still outperforms the random baseline and remains close to Glide at moderate recall levels. This indicates that most of the ranking power comes from broad priors learned across kinases in SMPBind I, while target specific examples mainly sharpen the very highest scoring region and improve the ordering of the hardest to recover binders.

\vspace{0.75em}

\begin{table}[h!]
\centering
\caption{AChE enrichment comparison. Each value shows the percentage of the ranked library required (ascending order) to recover $k$\% of all experimental actives, regardless of potency ranking. Lower values indicate better enrichment.}
\begin{adjustbox}{max width=\textwidth}
\begin{tabular}{lcccccc}
\toprule
\textbf{$k$\% AR} & \textbf{\# actives} & \textbf{Glide gscore} & \textbf{Boltz-2} & \textbf{Tensor-DTI} & \textbf{Random} \\
\midrule
1\%   & 4   & 0.53 & 0.53 & 0.53 & 1.00 \\
5\%   & 19  & 2.54 & 2.53 & 2.54 & 5.00 \\
20\%  & 75  & 10.01 & 13.45 & 11.21 & 20.00 \\
50\%  & 188 & 26.44 & 38.35 & 30.97 & 50.00 \\
100\% & 375 & 100.00 & 100.00 & 100.00 & 100.00 \\
\bottomrule
\end{tabular}
\end{adjustbox}
\label{tab:cdk2_enrichment}
\end{table}

\begin{table}[h!]
\centering
\caption{AChE enrichment comparison. Each value indicates the percentage of the ranked compound library required to recover the specified fraction of experimentally validated acetylcholinesterase binders. Lower percentages indicate earlier recovery (better enrichment). Tensor-DTI achieves the best performance across all cutoffs, confirming robust out-of-family generalization.}
\begin{adjustbox}{max width=\textwidth}
\begin{tabular}{lccccc}
\toprule
\textbf{Top-$k$} & \textbf{\# actives} & \textbf{Glide gscore} & \textbf{Boltz-2} & \textbf{Tensor-DTI} & \textbf{Random} \\
\midrule
1\%   & 4   & 86.25 & 39.07 & 37.60 & 80.10 \\
5\%   & 19  & 95.06 & 96.27 & 73.60 & 90.50 \\
20\%  & 75  & 99.20 & 99.07 & 73.60 & 95.50 \\
50\%  & 188 & 99.20 & 99.07 & 99.33 & 97.60 \\
100\% & 375 & 100.00 & 100.00 & 100.00 & 98.60 \\
\bottomrule
\end{tabular}
\end{adjustbox}
\label{tab:ache_enrichment}
\end{table}

For acetylcholinesterase, where all AChE and cholinesterase family interactions were removed from training, the three methods behave similarly at the lowest recall thresholds. To recover larger fractions of actives, Glide requires the smallest fraction of the library, with Tensor-DTI following closely and Boltz-2 lagging behind. In the top-k view Glide again retains the lowest recovery at small library fractions. Tensor-DTI therefore does not dominate on this non-kinase target but remains competitive with classical docking and clearly stronger than Boltz-2 once we move beyond the very first hits.

For human monoamine oxidase A, all oxidase family interactions were removed from training so the task probes generalization to a different structural and chemical regime. At the very lowest recall thresholds Boltz-2 slightly outperforms the other methods. Once we aim to recover more than a few percent of actives, Tensor-DTI becomes the most efficient option, reaching five to fifty percent of the active set after screening a smaller fraction of the library than either Glide or Boltz-2. In the top-k view docking and Boltz-2 maintain very low recovery within the strict top slices of the ranking, while Tensor-DTI offers a more favorable trade off between recall and screening budget as soon as one moves beyond the very first hits.

\begin{table}[h!]
\centering
\caption{MAO-A  enrichment comparison. Each value shows the percentage of the ranked library required (ascending order) to recover $k$\% of all experimental actives, regardless of potency ranking. Lower values indicate better enrichment.}
\begin{adjustbox}{max width=\textwidth}
\begin{tabular}{lcccccc}
\toprule
\textbf{$k$\% AR} & \textbf{\# actives} & \textbf{Glide gscore} & \textbf{Boltz-2} & \textbf{Tensor-DTI} & \textbf{Random} \\
\midrule
1\%   & 10   & 0.65 & 0.52 & 0.60 & 1.00 \\
5\%   & 50   & 4.50 & 2.80 & 2.65 & 5.00 \\
20\%  & 200  & 22.07 & 12.91 & 11.10 & 20.00 \\
50\%  & 499  & 55.16 & 37.86 & 31.35 & 50.00 \\
100\% & 998  & 100.00 & 100.00 & 99.80 & 100.00 \\
\bottomrule
\end{tabular}
\end{adjustbox}
\label{tab:cdk2_enrichment}
\end{table}

\begin{table}[h!]
\centering
\caption{MAO-A enrichment comparison. Each value indicates the percentage of the ranked library required to recover the specified fraction of top-affinity experimental binders (higher pIC50 preferred; tie-break by lower value). Lower values indicate better early recovery. Tensor-DTI was trained with all oxidase-family interactions removed.}
\vspace{0.15in}
\label{tab:maoA_enrichment}
\begin{adjustbox}{max width=\textwidth}
\begin{tabular}{lccccc}
\toprule
\textbf{Top-$k$} & \textbf{\# actives} & \textbf{Glide gscore} & \textbf{Boltz-2} & \textbf{Tensor-DTI} & \textbf{Random} \\
\midrule
1\%   & 10   & 97.95 & 100.00 & 63.10 & 90.90 \\
5\%   & 50   & 98.40 & 100.00 & 95.90 & 98.10 \\
20\%  & 200  & 98.40 & 100.00 & 96.50 & 99.60 \\
50\%  & 499  & 100.00 & 100.00 & 98.65 & 99.75 \\
100\% & 998  & 100.00 & 100.00 & 99.80 & 100.00 \\
\bottomrule
\end{tabular}
\end{adjustbox}
\end{table}

Taken together, these results outline a practical division of labor. Boltz-2 is extremely effective in its native setting of well parameterized ATP-competitive kinase pockets and excels when the goal is to find the earliest binders. Tensor-DTI offers complementary strengths. It achieves the most efficient global recovery of CDK2 actives, it remains competitive on AChE where Boltz-2 struggles, and it clearly improves the budget required to reach moderate recall on MAO-A compared with purely physics based scoring. Combined with the confidence and unfamiliarity diagnostics in Sections~\ref{confidence_model} and~\ref{cdk2_screening}, this positions Tensor-DTI as a robust partner to docking and Boltz-2 in large scale screening, both in terms of computational efficiency and especially when targets depart from the best studied kinase regime or when one cares about recovering more than only the very first hits.

\subsection{Broadening the Scope of Biomolecular Interaction Predictions}

Beyond small molecules, Tensor-DTI models peptide-protein, protein-RNA, and drug-RNA interactions, expanding its applicability to biologics and RNA therapeutics. For peptide-protein interactions, Tensor-DTI captures the physicochemical and sequence-dependent features governing peptide binding, achieving an AUPR of \(0.953 \pm 0.001\) on the Propedia \citep{propedia} dataset (Table~\ref{tab:tensor_dti_propedia} in Appendix \ref{biomolecular_pred}).

Similarly, for protein-RNA interactions, which are central to post-transcriptional regulation, the model achieves an AUPR of \(0.916 \pm 0.008\) on CoPRA \citep{CoPRA} (Table~\ref{tab:tensor_dti_copra} in Appendix \ref{biomolecular_pred}). When evaluated on PRA310, which provides affinity measurements for protein-RNA pairs, Tensor-DTI achieves a PCC of \(0.631 \pm 0.111\) for \(K_d\) (binding constant) and \(0.621 \pm 0.052\) for \(\Delta G\) (free energy), with corresponding RMSE values of \(1.443 \pm 0.232\) and \(1.910 \pm 0.212\) (Table~\ref{tab:protein_rna} in Appendix \ref{biomolecular_pred}). While a one-hot encoding baseline performed similarly in RMSE, Tensor-DTI exhibited stronger correlation with true affinities, indicating better predictive accuracy. 

For drug-RNA interactions, Tensor-DTI was trained on drug-RNA pairs from PDBBind, achieving a PCC of \(0.792 \pm 0.015\) and an RMSE of \(1.684 \pm 0.038\), outperforming the one-hot encoding baseline, which obtained a PCC of \(0.633 \pm 0.018\) and an RMSE of \(1.738 \pm 0.036\) (Table~\ref{tab:drug_rna} in Appendix \ref{biomolecular_pred}). Although RMSE values remained comparable, Tensor-DTI’s higher PCC suggests superior learning of structure-function relationships, capturing meaningful interaction patterns that conventional encoding methods fail to generalize.

These results demonstrate Tensor-DTI’s ability to generalize beyond small-molecule interactions, making it a versatile tool for modeling peptide and RNA interactions in therapeutic applications.

\section{Conclusion}

Accurate DTI prediction remains a challenge in computational drug discovery, requiring models that effectively capture the biochemical and structural determinants of molecular recognition. Tensor-DTI enhances DTI prediction by integrating multimodal embeddings from molecular graphs, protein language models, and binding site predictions within a contrastive learning framework. This multimodal design enables Tensor-DTI to improve predictive accuracy across diverse DTI benchmarks (BIOSNAP, BindingDB, and DAVIS) and to generalize to unseen drugs and proteins. In addition to binary interaction prediction, Tensor-DTI seamlessly extends to DTA regression, achieving competitive performance on challenging benchmarks such as TDC-DG and LP-PDBBind under strict domain generalization and low-leakage settings. The inclusion of contrastive learning objectives promotes the formation of robust, generalizable representations, as exemplified by the DUD-E benchmark, and further allows Tensor-DTI to improve performance on low-leak benchmarks such as PLINDER.

A key feature of Tensor-DTI is its explicit incorporation of pocket embeddings, which refine binding-site specificity and offer a structured, interpretable alternative to purely sequence-based or global structural embeddings in DTI models. By capturing both global and localized molecular features, the model enhances interaction modeling while maintaining flexibility for different molecular modalities. Nevertheless, our large-scale screening experiments indicate that the performance of pocket-conditioned Tensor-DTI variants is limited by the size and diversity of available pocket datasets, with the strongest reliability observed for PLINDER and selected cryptic-site systems.

The large-scale screening experiments further highlight Tensor-DTI’s capacity to generalize beyond its training domain. In the CDK2 screening, models trained with and without CDK2 produced qualitatively similar prediction patterns, recovering separable distributions between Tensor-DTI predicted active and inactive across Glide gscores, ligand efficiencies, and unfamiliarity. By filtering compounds to those with unfamiliarity below 1.0, representing the region of confident predictions, the model maintained chemically coherent and biologically meaningful predictions even without prior exposure to the target, albeit with some degradation relative to the in-domain setting. These observations suggest that Tensor-DTI captures transferable biochemical regularities rather than relying solely on target memorization, enabling reliable inference on related but unseen proteins. More generally, the combined use of confidence and unfamiliarity metrics provides a practical way to navigate the boundary between interpolation and extrapolation, helping ensure that predictions remain interpretable and trustworthy even at the frontier of chemical diversity. In the pocket scenario, while pocket-conditioned architectures provide valuable mechanistic interpretability, scaling them to large-scale screening will require substantially larger and more diverse datasets to achieve robust performance.

Consistent with this picture, our enrichment analysis shows Boltz-2 leading on CDK2 when the goal is to recover the very first binders in an ATP-competitive kinase pocket. Meanwhile, Tensor-DTI provides the most efficient full-recall ordering on CDK2, remains competitive with Glide on AChE, and substantially improves the screening budget required to reach moderate recall on MAO-A under family holdout, highlighting robust out-of-family generalization in these regimes.

Beyond small-molecule interactions, Tensor-DTI exhibits versatility in biomolecular interaction modeling, extending its applicability to peptide-protein and RNA-associated interactions. This broader scope makes Tensor-DTI particularly valuable for development of therapeutic agents beyond small molecules. 

Furthermore, its efficiency makes it suitable for large-scale virtual screening against ultra-large chemical libraries of billions of molecules, where conventional docking methods or diffusion models like Boltz-2 are computationally prohibitive. This scalability enables rapid hypothesis generation and prioritization at a scale that aligns with modern enumerated and on-demand chemical spaces. Moreover, the structured latent space could be further leveraged for generative modeling, enabling the exploration of new active compounds based on proximity in the learned representation space.

Overall, Tensor-DTI represents a scalable and generalizable framework for interaction modeling, balancing accuracy, interpretability, and computational efficiency. Future work will focus on refining its ability to model multi-target interactions, extending edge-of-domain calibration to novel protein classes, and integrating active learning strategies to further improve predictive robustness and real-world applicability.

\section{Methods}

\subsection{Model architecture}

\begin{figure*}[h!]
    \centering
    \includegraphics[width=0.8\textwidth]{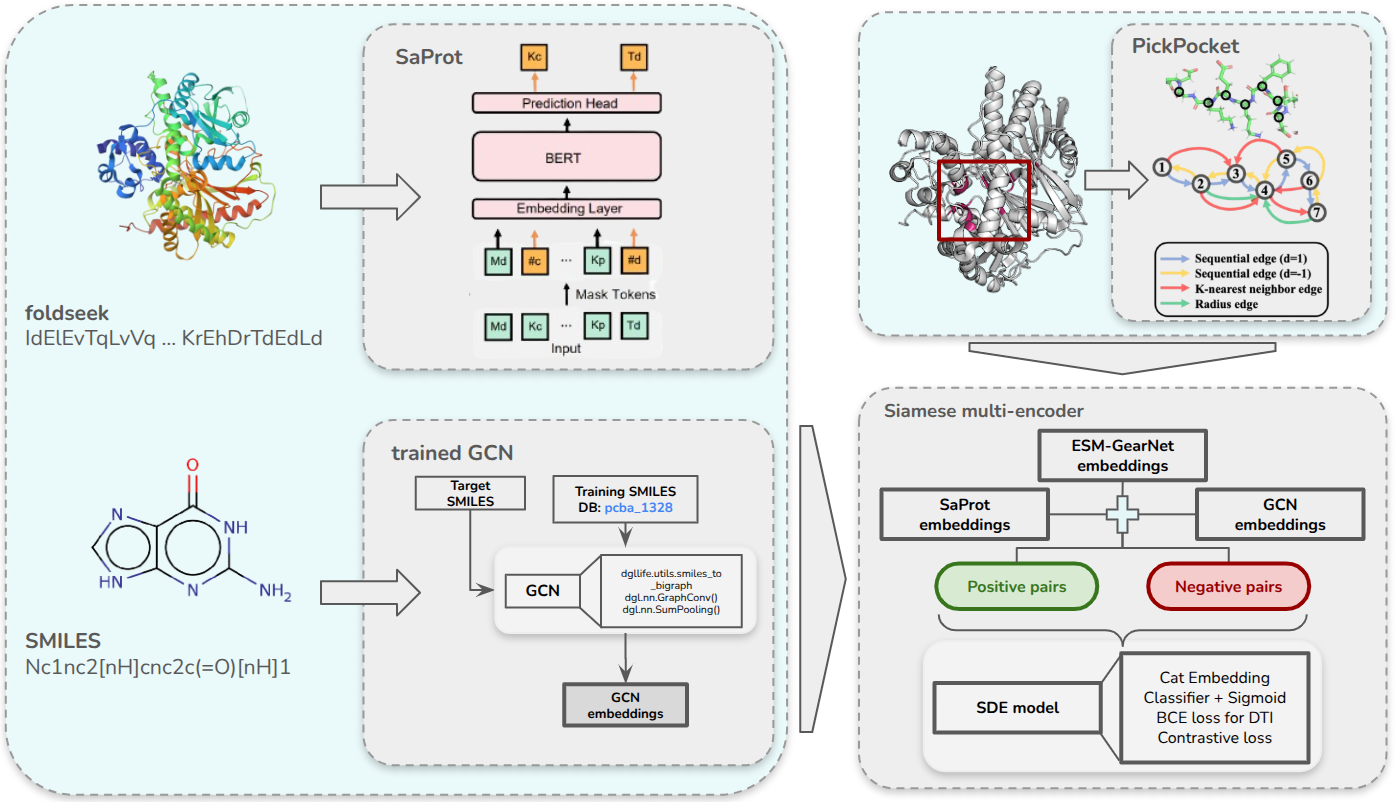}
   \caption{Tensor-DTI architecture with pocket embeddings. The model extends the base architecture by incorporating binding pocket representations, enabling site-specific interaction modeling. The protein shown is PDBID: 5ISX. The SaProt image is adapted from \citep{saprot}, and the Pickpocket image is adapted from \citep{gearnet}.}
    \label{fig:dti_pocket_architecture}
\end{figure*}

Tensor-DTI is a deep learning framework for DTI prediction that integrates multimodal molecular representations with contrastive learning. The model employs a siamese dual-encoder architecture \ref{fig:dti_pocket_architecture}, where separate encoder branches process drug and protein representations, projecting them into a shared latent space. A contrastive loss function encourages the embeddings of interacting pairs to cluster while pushing non-interacting pairs apart, enabling generalization to unseen drug-target pairs. For binary interaction classification, a binary cross-entropy loss is applied, ensuring the model learns a probabilistic interaction score. For affinity prediction, the model operates in a regression setting and is trained using a mean squared error loss to estimate continuous binding affinities.

Small molecules are represented as molecular graphs, and Tensor-DTI extracts drug embeddings using a Graph Convolutional Network (GCN) trained on PCBA\_1328, a dataset of 1.6M molecules with 1\,328 binary activity labels from PubChem \citep{pubchem}. The GCN iteratively aggregates local structural features, encoding bioactivity-relevant molecular patterns into a single graph-level embedding via sum pooling. Proteins are primarily represented using transformer-based embeddings from SaProt, a language model that integrates sequence and structural information. In the DTA scenario, we also used ESM-2, which encodes high-resolution sequence features from large-scale protein corpora. These embeddings provide rich contextualized representations of proteins, enabling the model to learn functionally relevant interaction patterns.

To refine binding-site specificity, Tensor-DTI incorporates pocket embeddings, allowing it to distinguish between global protein interactions and site-specific binding events. These embeddings are derived using PickPocket \citep{pickpocket}, which refines protein language model embeddings with structural information via GearNet \citep{gearnet}, a graph-based message-passing network that captures residue-residue interactions. Pocket embeddings are combined with full-protein representations, ensuring that the model effectively captures ligand-pocket interactions while maintaining contextual protein information. 

\subsection{Training procedure}

Specifically, we trained the DTI model with an addition of contrastive loss and binary cross-entropy (BCE) loss. The contrastive loss encourages positive drug-target pairs to cluster together in latent space while pushing negative pairs apart. Specifically, for a positive pair $(d,p)$ and corresponding negative pairs, we minimized:
\begin{align}
L_{\text{contrastive}} = &\sum_{(d,p)} \; \max \Big(0, \alpha + \|f_d(h_G) - f_p(h_P)\|_2 \nonumber \\
 \quad - &\|f_d(h_G) - f_p(h_P^-)\|_2 \Big),
\end{align}

where $\alpha$ is a margin hyperparameter, $h_P^-$ denotes a non-interacting (negative) protein embedding, $f_d(\cdot)$ is the projection head applied to the drug encoder output, and $f_p(\cdot)$ is the projection head applied to the protein encoder output.  The BCE loss is:
\begin{equation}
L_{\text{BCE}} = - \sum_{(d,p)} [y_{dp}\log(\hat{y}_{dp}) + (1 - y_{dp})\log(1 - \hat{y}_{dp})],
\end{equation}
where $y_{dp}$ is the ground-truth interaction label. We combined these losses to achieve robust, discriminative embeddings suited for both classification and interpretability.

We optimized the parameters using the Adam optimizer with a learning rate of $5 \times 10^{-5}$ for DTI tasks, weight decay of $1 \times 10^{-5}$, and early stopping based on validation performance. Multiple runs ensured statistical robustness, and final reported metrics were averaged across runs.

Tensor-DTI integrates pocket embeddings to refine binding-site specificity. These embeddings, derived using PickPocket \citep{pickpocket}, capture residue-residue interactions within functional binding sites. To combine protein and pocket representations, we apply a weighted aggregation:

\begin{equation}
\text{combined\_protein\_pocket} = \lambda_{protein} \cdot \text{encoded\_protein} + \lambda_{pocket} \cdot \text{encoded\_pocket}
\end{equation}

For all results reported in this study, we set $\lambda_{protein} = 1$ and $\lambda_{pocket} = 2$. This weighting emphasizes the binding site information while retaining global protein context. Further explanation and hyperparameter details are provided in Appendix~\ref{hyperparameters}.

A full visualization of the model architecture, including the integration of pocket embeddings, is provided in Appendix \ref{dti_visual}.

Additionally, Tensor-DTI includes an auxiliary confidence model that estimates the reliability of each predicted interaction. Although this component was not directly used in the benchmark evaluations, it plays a central role in prospective applications where experimental validation is limited. The confidence model assigns a reliability score to every prediction, allowing the prioritization of candidates with high certainty even in the absence of ground truth (see Appendix~\ref{confidence_model} for implementation details). Complementary to this, we integrate an unfamiliarity metric derived from a molecular autoencoder \citep{dl_edge}, which measures how far a compound lies from the model’s learned chemical manifold. Together, these two signals, confidence and unfamiliarity, provide an interpretable reliability framework that guides compound selection in large-scale inference and helps delineate the model’s operational boundary within chemical space.

\subsection{Adapting to affinity predictions}

For drug-target affinity (DTA) prediction, Tensor-DTI is adapted to a regression framework by replacing the contrastive and binary cross entropy losses with mean squared error loss. This modification allows the model to predict continuous affinity values instead of binary interactions. Drug and protein embeddings remain consistent with those used in classification tasks, ensuring a unified approach across predictive settings. The model is optimized using the Adam optimizer, with early stopping applied to prevent overfitting based on validation performance. By leveraging contrastive learning for classification and adapting seamlessly to affinity prediction, Tensor-DTI provides a flexible and scalable approach for modeling molecular interactions across diverse biological contexts. Details on the hyperparameters used for different settings can be found in Appendix \ref{hyperparameters}.

\subsection{Extending to other biomolecular representations}

Beyond small-molecule interactions, Tensor-DTI extends to RNA-protein and peptide-protein interactions, broadening its applicability to biomolecular modeling. Peptide representations are extracted from PeptideBERT \citep{peptidebert}, a transformer-based model trained on peptide sequences, while RNA embeddings are generated using ChaRNABERT \citep{charnabert}, which employs gradient-based subword tokenization to dynamically segment RNA sequences, capturing both nucleotide-level interactions and higher-order structural dependencies. These additional representations allow the model to extend beyond conventional drug-protein interactions and accommodate alternative therapeutic modalities.

\subsection{Glide Docking Protocol}\label{glide_methods}

Docking simulations were carried out using Schrödinger’s Extra Precision Glide (XP Glide) \citep{glide}. For each protein target, a docking grid was generated around the corresponding active site using a 10~Å inner box and a 30~Å outer box.

For acetylcholinesterase (AChE, PDB: 1C2B), the grid was centered on the catalytic triad (GLU334, HIS447, SER203). For monoamine oxidase A (MAO-A, PDB: 2BXR), the grid was centered on the catalytic site occupied by the co-crystallized ligand MGL. For cyclin-dependent kinase 2 (CDK2, PDB: 3BHV), the grid was centered on the binding site of the co-crystallized ligand VAR. Up to five docking poses were generated per compound, and the best pose was retained based on the Glide gscore. No positional constraints were applied for 1C2B and 2BXR, while a hydrogen-bond constraint to residue LEU83 was used for CDK2.

\section*{Acknowledgments}
This work was funded by project CPP2022-009737, financed by the Spanish Ministry of Science and Innovation (MICIU/AEI/10.13039/501100011033), and by the European Union (NextGenerationEU/PRTR).

\newpage

\bibliography{iclr2025_conference}

@article{moltrans,
    author = {Huang, Kexin and Xiao, Cao and Glass, Lucas M and Sun, Jimeng},
    title = {{MolTrans}: Molecular Interaction Transformer for drug–target interaction prediction},
    journal = {Bioinformatics},
    volume = {37},
    number = {6},
    pages = {830-836},
    year = {2020},
    month = {10},
}

@article{ConPlex,
  title={Contrastive learning in protein language space predicts interactions between drugs and protein targets},
  author={Singh, Rohit and Sledzieski, Samuel and Bryson, Bryan and Cowen, Lenore and Berger, Bonnie},
  journal={Proceedings of the National Academy of Sciences},
  volume={120},
  number={24},
  pages={e2220778120},
  year={2023},
  publisher={National Acad Sciences}
}

@article{leakproof,
  title={Leak proof {PDBBind}: A reorganized dataset of protein-ligand complexes for more generalizable binding affinity prediction},
  author={Li, Jie and Guan, Xingyi and Zhang, Oufan and Sun, Kunyang and Wang, Yingze and Bagni, Dorian and Head-Gordon, Teresa},
journal={arXiv preprint arXiv:2308.09639},
  year={2024}
}

@article{optipdbbind,
  title={{PDBBind} Optimization to Create a High-Quality Protein-Ligand Binding Dataset for Binding Affinity Prediction},
  author={Wang, Yingze and Sun, Kunyang and Li, Jie and Guan, Xingyi and Zhang, Oufan and Bagni, Dorian and Head-Gordon, Teresa},
  journal={arXiv preprint arXiv:2411.01223},
  year={2024}
}

@article{CoPRA,
      title={{CoPRA}: Bridging Cross-domain Pretrained Sequence Models with Complex Structures for Protein-{RNA} Binding Affinity Prediction}, 
      author={Rong Han and Xiaohong Liu and Tong Pan and Jing Xu and Xiaoyu Wang and Wuyang Lan and Zhenyu Li and Zixuan Wang and Jiangning Song and Guangyu Wang and Ting Chen},
      year={2024},
      journal={arXiv preprint arXiv:2409.03773}
}

@article{Lipinski,
title = {Experimental and computational approaches to estimate solubility and permeability in drug discovery and development settings},
journal = {Advanced Drug Delivery Reviews},
volume = {23},
number = {1},
pages = {3-25},
year = {1997},
author = {Christopher A. Lipinski and Franco Lombardo and Beryl W. Dominy and Paul J. Feeney},
}

@article{glide,
  title={Glide: a new approach for rapid, accurate docking and scoring. 1. {M}ethod and assessment of docking accuracy},
  author={Friesner, Richard A and Banks, Jay L and Murphy, Robert B and Halgren, Thomas A and Klicic, Jasna J and Mainz, Daniel T and Repasky, Matthew P and Knoll, Eric H and Shelley, Mee and Perry, Jason K and others},
  journal={Journal of Medicinal Chemistry},
  volume={47},
  number={7},
  pages={1739--1749},
  year={2004},
  publisher={ACS Publications}
}

@inproceedings{
diffdock,
title={{DiffDock}: Diffusion Steps, Twists, and Turns for Molecular Docking},
author={Gabriele Corso and Hannes St{\"a}rk and Bowen Jing and Regina Barzilay and Tommi S. Jaakkola},
booktitle={The Eleventh International Conference on Learning Representations },
year={2023},
}

@article{tankbind,
  title={{TANKBind}: Trigonometry-aware neural networks for drug-protein binding structure prediction},
  author={Lu, Wei and Wu, Qifeng and Zhang, Jixian and Rao, Jiahua and Li, Chengtao and Zheng, Shuangjia},
  journal={Advances in Neural Information Processing Systems},
  volume={35},
  pages={7236--7249},
  year={2022}
}

@article{chembl,
  title={{ChEMBL}: a large-scale bioactivity database for drug discovery},
  author={Gaulton, Anna and Bellis, Louisa J and Bento, A Patricia and Chambers, Jon and Davies, Mark and Hersey, Anne and Light, Yvonne and McGlinchey, Shaun and Michalovich, David and Al-Lazikani, Bissan and others},
  journal={Nucleic Acids Research},
  volume={40},
  number={D1},
  pages={D1100--D1107},
  year={2012},
  publisher={Oxford University Press}
}

@article{pdbbind,
  title={The {PDBbind} database: Collection of binding affinities for protein- ligand complexes with known three-dimensional structures},
  author={Wang, Renxiao and Fang, Xueliang and Lu, Yipin and Wang, Shaomeng},
  journal={Journal of Medicinal Chemistry},
  volume={47},
  number={12},
  pages={2977--2980},
  year={2004},
  publisher={ACS Publications}
}

@article{dude,
  title={Directory of useful decoys, enhanced ({DUD-E}): better ligands and decoys for better benchmarking},
  author={Mysinger, Michael M and Carchia, Michael and Irwin, John J and Shoichet, Brian K},
  journal={Journal of Medicinal Chemistry},
  volume={55},
  number={14},
  pages={6582--6594},
  year={2012},
  publisher={ACS Publications}
}

@article{esm,
  author={Rives, Alexander and Meier, Joshua and Sercu, Tom and Goyal, Siddharth and Lin, Zeming and Liu, Jason and Guo, Demi and Ott, Myle and Zitnick, C. Lawrence and Ma, Jerry and Fergus, Rob},
  title={Biological Structure and Function Emerge from Scaling Unsupervised Learning to 250 Million Protein Sequences},
  year={2019},
  journal={PNAS}
}

@article{saprot,
  title={{SaProt}: Protein language modeling with structure-aware vocabulary},
  author={Su, Jin and Han, Chenchen and Zhou, Yuyang and Shan, Junjie and Zhou, Xibin and Yuan, Fajie},
  journal={bioRxiv 2023.10.01.560349},
  year={2023},
  publisher={Cold Spring Harbor Laboratory}
}

@article{deepconv,
  title={{DeepConv-DTI}: Prediction of drug-target interactions via deep learning with convolution on protein sequences},
  author={Lee, Ingoo and Keum, Jongsoo and Nam, Hojung},
  journal={PLoS computational biology},
  volume={15},
  number={6},
  pages={e1007129},
  year={2019},
  publisher={Public Library of Science San Francisco, CA USA}
}

@article{transformercpi,
  title={{TransformerCPI}: improving compound--protein interaction prediction by sequence-based deep learning with self-attention mechanism and label reversal experiments},
  author={Chen, Lifan and Tan, Xiaoqin and Wang, Dingyan and Zhong, Feisheng and Liu, Xiaohong and Yang, Tianbiao and Luo, Xiaomin and Chen, Kaixian and Jiang, Hualiang and Zheng, Mingyue},
  journal={Bioinformatics},
  volume={36},
  number={16},
  pages={4406--4414},
  year={2020},
  publisher={Oxford University Press}
}

@article{hyperattention,
  title={{HyperAttentionDTI}: improving drug--protein interaction prediction by sequence-based deep learning with attention mechanism},
  author={Zhao, Qichang and Zhao, Haochen and Zheng, Kai and Wang, Jianxin},
  journal={Bioinformatics},
  volume={38},
  number={3},
  pages={655--662},
  year={2022},
  publisher={Oxford University Press}
}

@article{davis,
  title={Comprehensive analysis of kinase inhibitor selectivity},
  author={Davis, Mindy I and Hunt, Jeremy P and Herrgard, Sanna and Ciceri, Pietro and Wodicka, Lisa M and Pallares, Gabriel and Hocker, Michael and Treiber, Daniel K and Zarrinkar, Patrick P},
  journal={Nature Biotechnology},
  volume={29},
  number={11},
  pages={1046--1051},
  year={2011},
}

@article{graphdta,
  title={{GraphDTA}: predicting drug--target binding affinity with graph neural networks},
  author={Nguyen, Thin and Le, Hang and Quinn, Thomas P and Nguyen, Tri and Le, Thuc Duy and Venkatesh, Svetha},
  journal={Bioinformatics},
  volume={37},
  number={8},
  pages={1140--1147},
  year={2021},
  publisher={Oxford University Press}
}

@article{gnncti,
  title={Compound--protein interaction prediction with end-to-end learning of neural networks for graphs and sequences},
  author={Tsubaki, Masashi and Tomii, Kentaro and Sese, Jun},
  journal={Bioinformatics},
  volume={35},
  number={2},
  pages={309--318},
  year={2019},
  publisher={Oxford University Press}
}

@article{pocketdta,
  title={{PocketDTA}: an advanced multimodal architecture for enhanced prediction of drug- target affinity from {3D} structural data of target binding pockets},
  author={Zhao, Long and Wang, Hongmei and Shi, Shaoping},
  journal={Bioinformatics},
  volume={40},
  number={10},
  pages={btae594},
  year={2024},
  publisher={Oxford University Press}
}

@article {plinder,
	author = {Durairaj, Janani and Adeshina, Yusuf and Cao, Zhonglin and Zhang, Xuejin and Oleinikovas, Vladas and Duignan, Thomas and McClure, Zachary and Robin, Xavier and Studer, Gabriel and Kovtun, Daniel and Rossi, Emanuele and Zhou, Guoqing and Veccham, Srimukh and Isert, Clemens and Peng, Yuxing and Sundareson, Prabindh and Akdel, Mehmet and Corso, Gabriele and St{\"a}rk, Hannes and Tauriello, Gerardo and Carpenter, Zachary and Bronstein, Michael and Kucukbenli, Emine and Schwede, Torsten and Naef, Luca},
	title = {{PLINDER}: The protein-ligand interactions dataset and evaluation resource},
	elocation-id = {2024.07.17.603955},
	year = {2024},
	publisher = {Cold Spring Harbor Laboratory},
	journal = {bioRxiv 2024.07.17.603955}

}

@article{propedia,
  title={Propedia v2. 3: A novel representation approach for the peptide-protein interaction database using graph-based structural signatures},
  author={Martins, Pedro and Mariano, Diego and Carvalho, Frederico Chaves and Bastos, Luana Luiza and Moraes, Lucas and Paix{\~a}o, Vivian and Cardoso de Melo-Minardi, Raquel},
  journal={Frontiers in Bioinformatics},
  volume={3},
  pages={1103103},
  year={2023},
  publisher={Frontiers Media SA}
}

@article{esm2,
  title={Evolutionary-scale prediction of atomic-level protein structure with a language model},
  author={Lin, Zeming and Akin, Halil and Rao, Roshan and Hie, Brian and Zhu, Zhongkai and Lu, Wenting and Smetanin, Nikita and Verkuil, Robert and Kabeli, Ori and Shmueli, Yaniv and others},
  journal={Science},
  volume={379},
  number={6637},
  pages={1123--1130},
  year={2023},
  publisher={American Association for the Advancement of Science}
}

@article{vaswani2017attention,
  title={Attention is all you need},
  author={Vaswani, Ashish and Shazeer, Noam and Parmar, Niki and Uszkoreit, Jakob and Jones, Llion and Gomez, Aidan N and Kaiser, {\L}ukasz and Polosukhin, Illia},
  journal={Advances in Neural Information Processing Systems},
  volume={30},
  year={2017}
}

@inproceedings{radford2021learning,
  title={Learning transferable visual models from natural language supervision},
  author={Radford, Alec and Kim, Jong Wook and Hallacy, Chris and Ramesh, Aditya and Goh, Gabriel and Agarwal, Sandhini and Sastry, Girish and Askell, Amanda and Mishkin, Pamela and Clark, Jack and others},
  booktitle={International Conference on Machine Learning},
  pages={8748--8763},
  year={2021},
  organization={PMLR}
}

@inproceedings{zbontar2021barlow,
  title={Barlow twins: Self-supervised learning via redundancy reduction},
  author={Zbontar, Jure and Jing, Li and Misra, Ishan and LeCun, Yann and Deny, St{\'e}phane},
  booktitle={International Conference on Machine Learning},
  pages={12310--12320},
  year={2021},
  organization={PMLR}
}

@article{finkelshtein2023cooperative,
  title={Cooperative graph neural networks},
  author={Finkelshtein, Ben and Huang, Xingyue and Bronstein, Michael and Ceylan, {\.I}smail {\.I}lkan},
  journal={arXiv preprint arXiv:2310.01267},
  year={2023}
}

@article{kipf2016semi,
  title={Semi-supervised classification with graph convolutional networks},
  author={Kipf, Thomas N and Welling, Max},
  journal={arXiv preprint arXiv:1609.02907},
  year={2016}
}

@article{mfps,
  title={Extended-connectivity fingerprints},
  author={Rogers, David and Hahn, Mathew},
  journal={Journal of Chemical Information and Modeling},
  volume={50},
  number={5},
  pages={742--754},
  year={2010},
  publisher={ACS Publications}
}

@article{charnabert,
  title={Character-level Tokenizations as Powerful Inductive Biases for {RNA} Foundational Models},
  author={Morales-Pastor, Adri{\'a}n and V{\'a}zquez-Reza, Raquel and Wiecz{\'o}r, Mi{\l}osz and Valverde, Cl{\`a}udia and Gil-Sorribes, Manel and Miquel-Oliver, Bertran and Ciudad, {\'A}lvaro and Molina, Alexis},
  journal={arXiv preprint arXiv:2411.11808},
  year={2024}
}

@article{peptidebert,
  title={{PeptideBERT}: A language model based on transformers for peptide property prediction},
  author={Guntuboina, Chakradhar and Das, Adrita and Mollaei, Parisa and Kim, Seongwon and Barati Farimani, Amir},
  journal={The Journal of Physical Chemistry Letters},
  volume={14},
  number={46},
  pages={10427--10434},
  year={2023},
  publisher={ACS Publications}
}

@article{apdomain,
  title={A systematic prediction of multiple drug-target interactions from chemical, genomic, and pharmacological data},
  author={Yu, Hua and Chen, Jianxin and Xu, Xue and Li, Yan and Zhao, Huihui and Fang, Yupeng and Li, Xiuxiu and Zhou, Wei and Wang, Wei and Wang, Yonghua},
  journal={PloS one},
  volume={7},
  number={5},
  pages={e37608},
  year={2012},
  publisher={Public Library of Science San Francisco, USA}
}

@inproceedings{
gearnet,
title={Protein Representation Learning by Geometric Structure Pretraining},
author={Zuobai Zhang and Minghao Xu and Arian Rokkum Jamasb and Vijil Chenthamarakshan and Aurelie Lozano and Payel Das and Jian Tang},
booktitle={The Eleventh International Conference on Learning Representations },
year={2023},
}

@article{cryptic_site,
title = {Grasping cryptic binding sites to neutralize drug resistance in the field of anticancer},
journal = {Drug Discovery Today},
volume = {28},
number = {9},
pages = {103705},
year = {2023},
author = {Wei-Cheng Yang and Dao-Hong Gong and  {Hong Wu} and Yang-Yang Gao and Ge-Fei Hao},
}

@article{enzpred_cti,
    author = {Goldman, Samuel AND Das, Ria AND Yang, Kevin K. AND Coley, Connor W.},
    journal = {PLOS Computational Biology},
    publisher = {Public Library of Science},
    title = {Machine learning modeling of family wide enzyme-substrate specificity screens},
    year = {2022},
    month = {02},
    volume = {18},
    pages = {1-20},
    number = {2},
}

@article{pubchem,
  title={{PubChem} 2023 update},
  author={Kim, Sunghwan and Chen, Jie and Cheng, Tiejun and Gindulyte, Asta and He, Jia and He, Siqian and Li, Qingliang and Shoemaker, Benjamin A and Thiessen, Paul A and Yu, Bo and others},
  journal={Nucleic Acids Research},
  volume={51},
  number={D1},
  pages={D1373--D1380},
  year={2023},
  publisher={Oxford University Press}
}

@article{zinc22,
author={Tingle, Benjamin I.
and Tang, Khanh G.
and Castanon, Mar
and Gutierrez, John J.
and Khurelbaatar, Munkhzul
and Dandarchuluun, Chinzorig
and Moroz, Yurii S.
and Irwin, John J.},
title={{ZINC-22} {A} Free Multi-Billion-Scale Database of Tangible Compounds for Ligand Discovery},
journal={Journal of Chemical Information and Modeling},
year={2023},
month={Feb},
day={27},
publisher={American Chemical Society},
volume={63},
number={4},
pages={1166-1176},
}

@misc{enamine,
  author = {Enamine},
  title = {{Enamine REAL Space}},
  howpublished = {\url{https://enamine.net/compound-collections/real-compounds/real-space-navigator}},
  note = {Accessed on November 01, 2024}
}

@inproceedings{
tdc,
title={Therapeutics Data Commons: Machine Learning Datasets and Tasks for Drug Discovery and Development},
author={Kexin Huang and Tianfan Fu and Wenhao Gao and Yue Zhao and Yusuf H Roohani and Jure Leskovec and Connor W. Coley and Cao Xiao and Jimeng Sun and Marinka Zitnik},
booktitle={Thirty-fifth Conference on Neural Information Processing Systems Datasets and Benchmarks Track (Round 1)},
year={2021}
}

@article{chemicalspace,
  title={Chemical space: limits, evolution and modelling of an object bigger than our universal library},
  author={Restrepo, Guillermo},
  journal={Digital Discovery},
  volume={1},
  number={5},
  pages={568--585},
  year={2022},
  publisher={Royal Society of Chemistry}
}

@article{cdk2therapies,
  title={CDK/cyclin dependencies define extreme cancer cell-cycle heterogeneity and collateral vulnerabilities},
  author={Knudsen, Erik S and Kumarasamy, Vishnu and Nambiar, Ram and Pearson, Joel D and Vail, Paris and Rosenheck, Hanna and Wang, Jianxin and Eng, Kevin and Bremner, Rod and Schramek, Daniel and others},
  journal={Cell reports},
  volume={38},
  number={9},
  year={2022},
  publisher={Elsevier}
}

@article{rettherapies,
  title={RET kinase alterations in targeted cancer therapy},
  author={Liu, Xuan and Hu, Xueqing and Shen, Tao and Li, Qi and Mooers, Blaine HM and Wu, Jie},
  journal={Cancer Drug Resistance},
  volume={3},
  number={3},
  pages={472},
  year={2020},
  publisher={OAE Publishing Inc}
}

@article{dl_edge,
  author    = {van Tilborg, Derek and Rossen, Lars and Grisoni, Fabio},
  title     = {Molecular deep learning at the edge of chemical space},
  journal   = {ChemRxiv},
  year      = {2025},
  note      = {Preprint, not peer-reviewed},
  doi       = {10.26434/chemrxiv-2025-qj4k3}
}

@article{pickpocket,
  author    = {Tarasi, Stelina and Malo, Laura and Molina, Alexis},
  title     = {Evolutionary and geometric signatures reveal ligand-binding sites across proteomes},
  journal   = {bioRxiv},
  year      = {2025},
  note      = {Preprint, not peer-reviewed},
  doi       = {10.1101/2025.10.07.680847}
}

@article{boltz2,
  author = {Passaro, Saro and Corso, Gabriele and Wohlwend, Jeremy and Reveiz, Mateo and Thaler, Stephan and Somnath, Vignesh Ram and Getz, Noah and Portnoi, Tally and Roy, Julien and Stark, Hannes and Kwabi-Addo, David and Beaini, Dominique and Jaakkola, Tommi and Barzilay, Regina},
  title = {Boltz-2: Towards Accurate and Efficient Binding Affinity Prediction},
  year = {2025},
  doi = {10.1101/2025.06.14.659707},
  journal = {bioRxiv}
}

@Article{surr,
author={McNutt, Andrew T.
and Francoeur, Paul
and Aggarwal, Rishal
and Masuda, Tomohide
and Meli, Rocco
and Ragoza, Matthew
and Sunseri, Jocelyn
and Koes, David Ryan},
title={GNINA 1.0: molecular docking with deep learning},
journal={Journal of Cheminformatics},
year={2021},
month={Jun},
day={09},
volume={13},
number={1},
pages={43},
issn={1758-2946},
doi={10.1186/s13321-021-00522-2},
url={https://doi.org/10.1186/s13321-021-00522-2}
}

@article{al,
  title     = {Accelerating high-throughput virtual screening through molecular pool-based active learning},
  author    = {Graff, David E and Shakhnovich, Eugene I and Coley, Connor W},
  journal   = {Chemical Science},
  volume    = {12},
  number    = {22},
  pages     = {7866--7881},
  year      = {2021},
  publisher = {Royal Society of Chemistry},
  doi       = {10.1039/d0sc06805e},
  pmid      = {34168840},
  pmcid     = {PMC8188596}
}

@article{sf,
      title={Scoreformer: A Surrogate Model For Large-Scale Prediction of Docking Scores}, 
      author={Álvaro Ciudad and Adrián Morales-Pastor and Laura Malo and Isaac Filella-Mercè and Victor Guallar and Alexis Molina},
      year={2024},
      eprint={2406.09346},
      archivePrefix={arXiv},
      primaryClass={cs.LG} 
}
\bibliographystyle{iclr2025_conference}

\newpage
\appendix

\section{DTI Model Visualization}\label{dti_visual}

\begin{figure*}[h!]
    \centering
    \includegraphics[width=\textwidth]{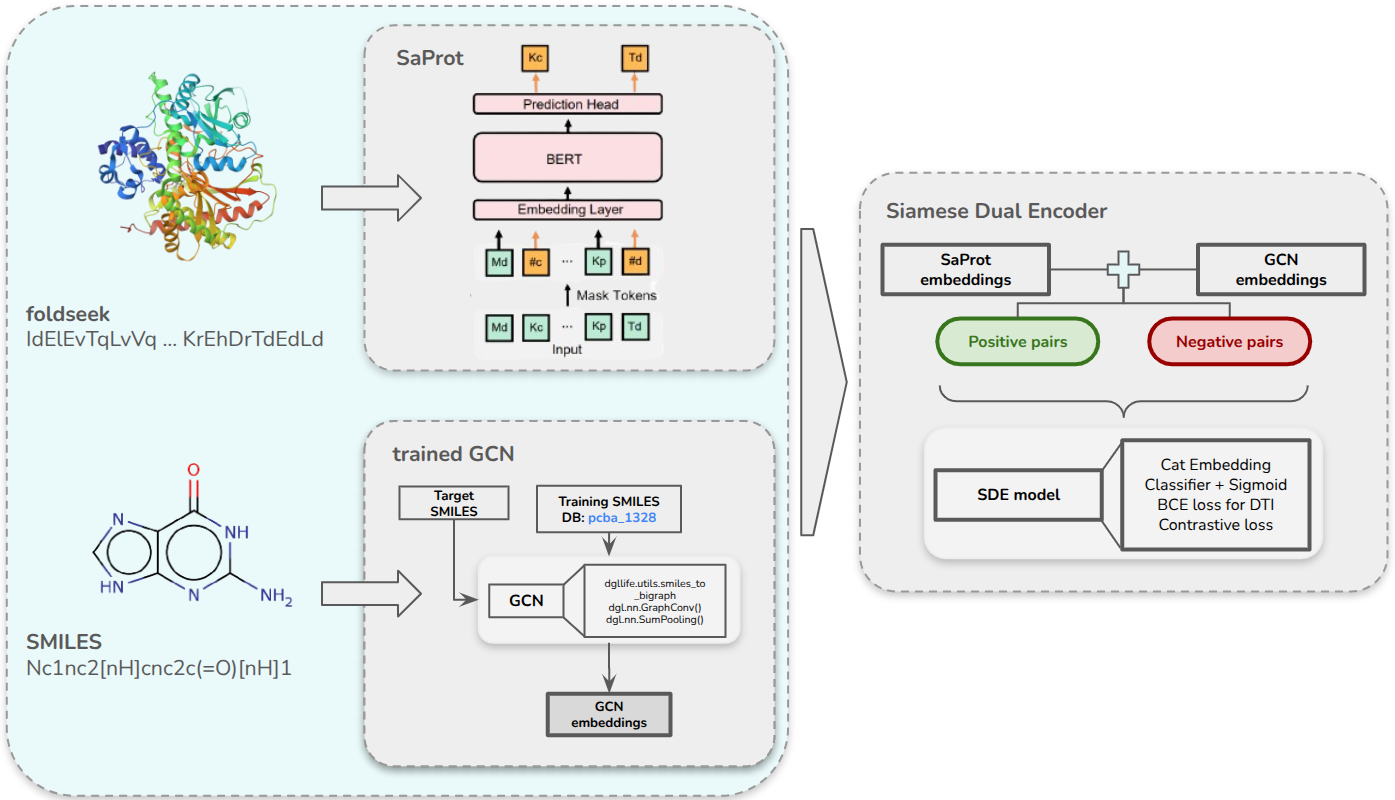}
    \caption{Tensor-DTI architecture. A siamese dual-encoder processes multimodal embeddings from drugs and proteins, using contrastive learning to refine the interaction space. The protein shown is 5ISX. The SaProt image is adapted from \citep{saprot}.}
    \label{fig:dti_architecture}
\end{figure*}

\begin{figure*}[h!]
    \centering
    \includegraphics[width=\textwidth]{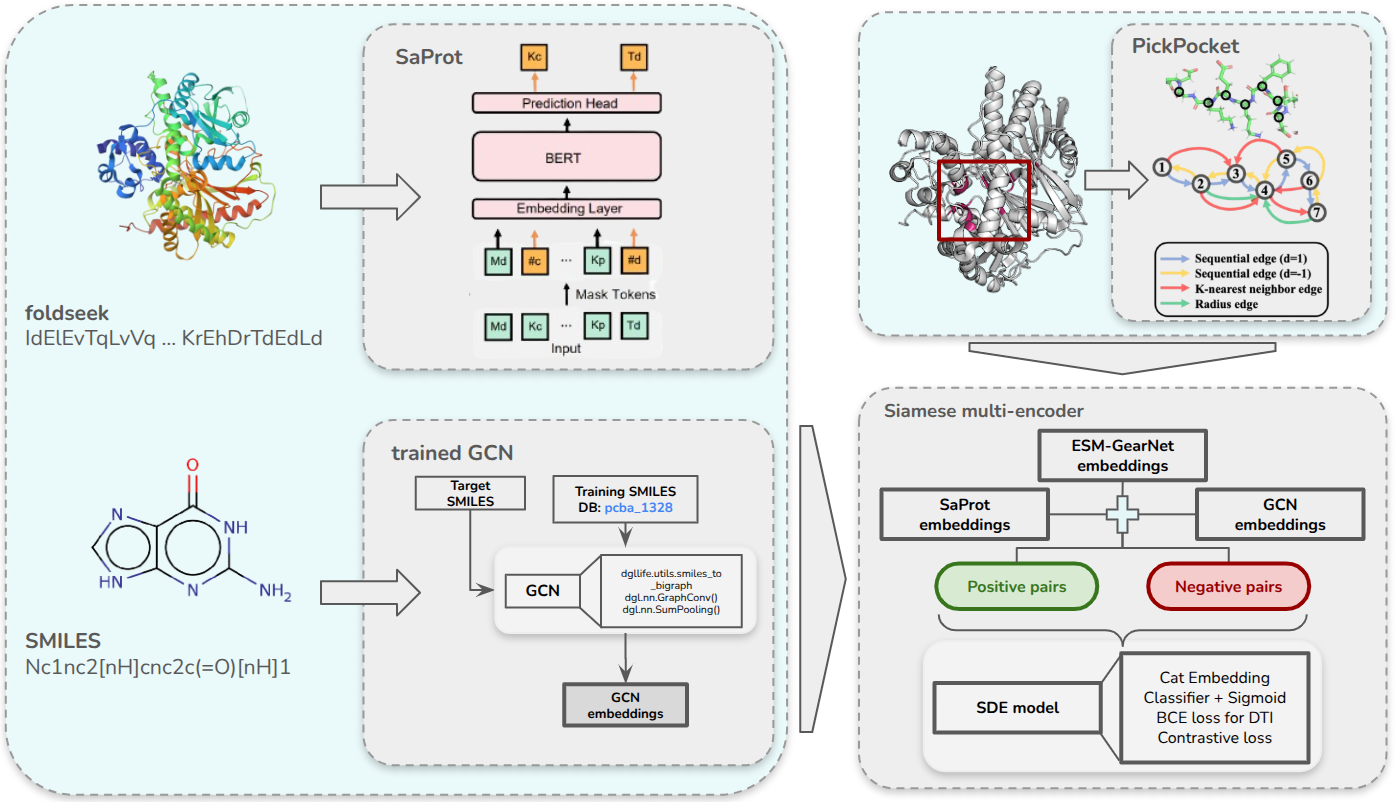}
   \caption{Tensor-DTI architecture with pocket embeddings. The model extends the base architecture by incorporating binding pocket representations, enabling site-specific interaction modeling. The protein shown is 5ISX. The SaProt image is adapted from \citep{saprot}, and the PickPocket image is adapted from \citep{gearnet}.}
    \label{fig:dti_pocket_architecture_app}
\end{figure*}

\newpage
\section{Hyperparameter Configurations and Model Architectures}\label{hyperparameters}

The choice of $\lambda_{\text{protein}} = 1$ and $\lambda_{\text{pocket}} = 2$ in:

\begin{equation}
h_{\text{protein}}^{*} = 
\lambda_{\text{protein}} \cdot h_{\text{protein}} + 
\lambda_{\text{pocket}} \cdot h_{\text{pocket}},
\end{equation}

was informed by validation performance on PLINDER. In an initial and more challenging imbalanced setting, this weighting combination achieved the highest AUPR and F1 scores:

\begin{table*}[h]
\centering
\caption{Performance of different $\alpha$, $\beta$ combinations on PLINDER.}
\vspace{0.15in}
\label{tab:plinder_ablation}

\begin{tabular}{cccc}
\toprule
\textbf{$\lambda_{protein}$} & \textbf{$\lambda_{pocket}$} & \textbf{AUPR} & \textbf{F1} \\
\midrule
0 & 1 & 0.633 & 0.405 \\
1 & 1 & 0.689 & 0.500 \\
\textbf{1} & \textbf{2} & \textbf{0.700} & \textbf{0.524} \\
1 & 3 & 0.696 & 0.507 \\
1 & 5 & 0.694 & 0.513 \\
1 & 7 & 0.688 & 0.515 \\
1 & 10 & 0.690 & 0.513 \\
\bottomrule
\end{tabular}
\end{table*}

\begin{table*}[h]
\centering
\caption{Hyperparameters used for DTI models in protein-drug interaction benchmarks.}
\vspace{0.15in}
\label{tab:dti_protein_drug}

\begin{tabular}{lccc}
\toprule
\textbf{Benchmark} & \textbf{Emb. Dim} & \textbf{Hidden Dim} & \textbf{Output Dim} \\
\midrule
BIOSNAP, BindingDB, DAVIS & (64, 1\,280) & 512 & 256 \\
DUD-E & (64, 1\,280) & 512 & 256 \\
SMPBind-I & (64, 1\,280) & 1\,024 & 512 \\
PLINDER (No Pocket) & (64, 1\,280) & 512 & 256 \\
\bottomrule
\end{tabular}

\vspace{0.2in} 

\begin{tabular}{lcc}  
\toprule
\textbf{Benchmark} & \textbf{LR} & \textbf{Epochs} \\
\midrule
BIOSNAP, BindingDB, DAVIS & 0.00005 & 1\,000 \\
DUD-E & 0.000005 & 300 \\
SMPBind-I & 0.00001 & 100 \\
PLINDER (No Pocket) & 0.00001 & 1\,000 \\
\bottomrule
\end{tabular}

\end{table*}

\begin{table*}[h!]
\centering
\caption{Hyperparameters used for DTI models in alternative biomolecular interaction benchmarks, including RNA, peptides, and pocket embeddings.}
\vspace{0.15in}
\label{tab:dti_other_biomolecules}

\begin{tabular}{lccc}
\toprule
\textbf{Benchmark} & \textbf{Emb. Dim} & \textbf{Hidden Dim} & \textbf{Out Dim} \\
\midrule
PLINDER (With Pocket) & (64, 1\,280, 1\,536) & 512 & 256 \\
CoPRA (Protein-RNA) & (480, 1\,280) & 512 & 256 \\
Propedia (Peptide-Protein) & (480, 1\,280) & 512 & 256 \\
\bottomrule
\end{tabular}

\vspace{0.2in} 

\begin{tabular}{lcc}
\toprule
\textbf{Benchmark} & \textbf{LR} & \textbf{Epochs} \\
\midrule
PLINDER (With Pocket) & 0.00001 & 1\,000 \\
CoPRA (Protein-RNA) & 0.00001 & 1\,000 \\
Propedia (Peptide-Protein) & 0.00001 & 1\,000 \\
\bottomrule
\end{tabular}

\end{table*}

\newpage

\subsection{Enrichment computation}\label{enrichment}

For each target, we compared three ranking criteria: Glide gscore \citep{glide}, Boltz-2, and Tensor-DTI, by how quickly they recover known binders. Given a candidate set of size $N$ with $A$ experimentally validated actives, we report cumulative recall at top-$k$ operating points corresponding to 1\%, 5\%, 20\%, 50\%, and 100\% of the ranked list. Recall@$k$ is a monotonic proxy for EF@$k$, since
\[
\mathrm{EF@}k \;=\; \frac{\mathrm{TP@}k/k}{A/N} \;=\; \mathrm{Recall@}k \cdot \frac{N}{k},
\]
and $N,k$ are fixed per comparison. Thus, higher recall at a given list percentage implies higher EF@$k$.

\paragraph{Ranking criteria.}
Glide gscore ranks compounds by Glide gscore in ascending order (more favorable scores first). Boltz-2 ranks by predicted affinity in ascending order. Tensor-DTI applies a two-key sort: (i) predicted label (positives first), and (ii) within each label, a confidence-based tie-breaker that prioritizes more certain predictions among positives (lower confidence score indicates higher certainty in our calibration; see Appendix~\ref{confidence_model}).

\paragraph{Alignment across methods.}
To ensure a fair comparison, we intersect the set of experimentally validated binders shared by all methods and compute recall with respect to this common active set.

\paragraph{Library sizes.}

For CDK2, the evaluated library contained 2\,450 compounds: 796 experimentally validated binders and 1\,654 random molecules sampled from ChEMBL.  
For AChE, the evaluated set comprised 750 molecules: 375 experimental binders and 375 random decoys selected from the Enamine REAL library. For MAO-A, the evaluated library comprised 1\,998 molecules: 998 experimental binders and 1\,000 random molecules selected from the Enamine REAL library.
 
\subsection{DTA} \label{hyper_s_dta}

\begin{table*}[h!]
\centering
\caption{Hyperparameters used for DTA models.}
\vspace{0.15in}
\label{tab:dta_hyperparameters}

\begin{tabular}{lccc}
\toprule
\textbf{Benchmark} & \textbf{Emb. Dim} & \textbf{Hidden Dim} & \textbf{Output Dim} \\
\midrule
TDC-DG (Molecule-Protein) & (2\,048, 1\,280) & 4096 & 1024 \\
LP-PDBBind (Leakproof) & (64, 1\,280) & 4\,096 & 1\,024 \\
PDBBind (Molecule-Protein) & (64, 1\,280) & 4\,096 & 1\,024 \\
PDBBind (Peptide-Protein) & (480, 1\,280) & 4\,096 & 1\,024 \\
PDBBind (RNA-Drug) & (480, 64) & 4\,096 & 1\,024 \\
\bottomrule
\end{tabular}

\vspace{0.2in} 

\begin{tabular}{lcc}
\toprule
\textbf{Benchmark} & \textbf{LR} & \textbf{Epochs} \\
\midrule
TDC-DG (Molecule-Protein) & 0.0001 & 1\,000 \\
LP-PDBBind (Leakproof) & 0.0001 & 1\,000 \\
PDBBind (Molecule-Protein) & 0.0001 & 1\,000 \\
PDBBind (Peptide-Protein) & 0.0001 & 1\,000 \\
PDBBind (RNA-Drug) & 0.0001 & 200 \\
\bottomrule
\end{tabular}

\end{table*}

\section{Confidence and Unfamiliarity Models}\label{confidence_model}

To ensure that Tensor-DTI predictions are both accurate and interpretable, we introduce two complementary mechanisms for assessing reliability: a Confidence Model that estimates prediction certainty, and an Unfamiliarity Model that evaluates whether a compound lies within the model’s learned chemical domain. Both models are trained jointly with Tensor-DTI using the SMPBind-I dataset, providing exposure to a broad range of chemical scaffolds and interaction patterns. The same framework is also employed for the PLINDER variant with pocket embeddings.

\subsection{Confidence Model}

To evaluate the reliability of its predictions, we introduce a Confidence Model, which was trained jointly with the primary Tensor-DTI model. The model processed the fused drug-target embeddings together with their interaction logits, estimating the certainty of each prediction through a continuous confidence score.

The confidence model was implemented as a feedforward neural network $f_{\text{conf}}$ that takes as input the concatenated drug-target embeddings and interaction logits, producing a single confidence score:

\[
c = f_{\text{conf}}(E_{\text{combined}}, \hat{y}),
\]

where \(E_{\text{combined}}\) represents the joint embedding of the drug-target pair, and \(\hat{y}\) is the predicted interaction score.

The confidence score was designed to approximate the absolute deviation between the predicted interaction probability and the ground truth:

\[
L_{\text{Conf}} = \frac{1}{N} \sum_{i} \left( c_i - \lvert y_i - \hat{y}_i \rvert \right)^2.
\]

Lower confidence values correspond to more reliable predictions, while higher scores indicate uncertainty or potential misclassification.  

\begin{figure}[h!]
    \centering
    \includegraphics[width=0.6\linewidth]{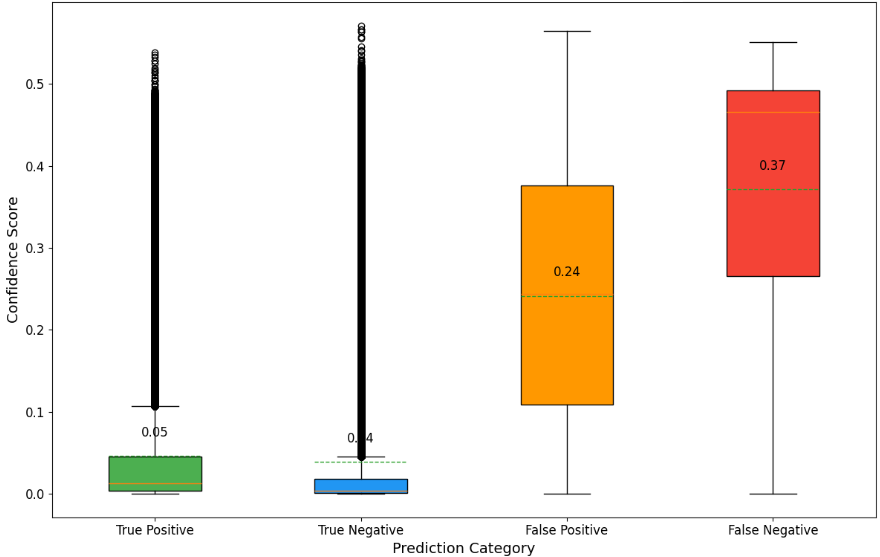}
    \caption{Distribution of confidence scores across prediction categories (TP, FP, TN, FN). Lower confidence values denote higher certainty, while higher scores indicate uncertainty or potential misclassification.}
    \label{fig:confidence_scores}
\end{figure}

After training, the model outputs a confidence score for each prediction, quantifying its reliability. To analyze how confidence correlates with prediction accuracy, predictions are categorized into True Positives (TP), False Positives (FP), True Negatives (TN), and False Negatives (FN).  
As illustrated in Figure~\ref{fig:confidence_scores}, correctly classified samples (TP and TN) exhibit lower confidence scores, indicating high certainty, whereas misclassified samples (FP and FN) tend to show higher scores, reflecting uncertainty in their predictions.

This confidence-aware framework enhances interpretability and enables systematic prioritization of high-confidence interactions for downstream experimental validation. Moreover, confidence values serve as a ranking metric, ensuring that selected top-scoring drug-target pairs are not only predicted as interacting but are also assigned high reliability (low confidence score).

\subsection{Unfamiliarity Model}

Complementary to confidence estimation, we introduce an Unfamiliarity Model to assess whether a compound lies within the chemical domain learned by the model. Following the framework of \citet{dl_edge}, this metric quantifies how “in-distribution" a molecule is with respect to the training chemical space.

We compute unfamiliarity via a jointly trained drug autoencoder that reconstructs SMILES from drug embeddings. The autoencoder encodes each drug into a latent space \(z\) and decodes a token sequence, trained with token-level cross-entropy:
\[
L_{\text{Recon}} \;=\; -\sum_{t} \log p(s_t \mid z),
\]
where \(s_t\) are SMILES tokens. For each molecule, we derive an unfamiliarity score as the (log) normalized negative log-likelihood (NLL) of the reconstruction:
\[
U \;=\; \log\!\big(\text{NLL}_{\text{recon}} + \epsilon\big).
\]
Higher \(U\) indicates the compound is farther from the model’s learned chemical manifold (out-of-distribution), while lower \(U\) denotes chemically familiar regions.

Empirically, correctly predicted interactions tend to show lower \(U\), whereas errors (false positives/negatives) concentrate at higher \(U\), consistent with distributional shift (as shown in \citep{dl_edge}). During large-scale screening, we restrict analysis to compounds with \(U < 1.0\), which delineates a chemically familiar regime where predictions remain more reliable.

Together, confidence and unfamiliarity provide complementary reliability signals. Confidence quantifies certainty about a given prediction, while unfamiliarity indicates whether that prediction was made within the model’s domain of competence. This dual criterion improves prioritization for prospective selection and downstream validation.

\subsection{Training Objective}

In practice, we optimize a composite objective that mirrors our implementation, combining interaction classification, representation separation, confidence calibration, and SMILES reconstruction. Let $\hat{y}$ be the interaction \emph{logit} (pre-sigmoid) produced by the classifier, and $\sigma(\hat{y})$ its probability. The total loss is
\[
L_{\text{Total}}
\;=\;
\alpha_{\text{cls}} \, L_{\text{BCE}}
\;+\;
\alpha_{\text{con}} \, L_{\text{Contrastive}}
\;+\;
\alpha_{\text{conf}} \, L_{\text{Conf}}
\;+\;
\alpha_{\text{recon}} \, L_{\text{Recon}}.
\]
\textbf{Classification.} We use binary cross-entropy with logits:
\[
L_{\text{BCE}}
= - \big[
y \log \sigma(\hat{y}) + (1-y)\log(1-\sigma(\hat{y}))
\big].
\]
\textbf{Contrastive separation.} We encourage embedding proximity for positives and separation for negatives using a cosine-distance margin loss:
\[
L_{\text{Contrastive}}
= \mathbb{E}\big[
y \, d^2 + (1-y)\,\max(0, m - d)^2
\big],\quad
d = 1 - \cos(\mathbf{e}_d,\mathbf{e}_p),
\]
where $\mathbf{e}_d,\mathbf{e}_p$ are the drug/protein embeddings and $m$ is a margin (set to $1.0$).

\textbf{Confidence calibration.} The confidence head receives the concatenated pair features (drug-target embedding and interaction logit) and is trained to predict the absolute error of the classifier:
\[
c = f_{\text{conf}}([\mathbf{e}_d \Vert \mathbf{e}_p], \hat{y}),\qquad
L_{\text{Conf}} = \big(c - \lvert y - \sigma(\hat{y}) \rvert\big)^2.
\]
By design, lower $c$ denotes higher certainty (smaller absolute error), matching our ranking convention.

\textbf{SMILES reconstruction (unfamiliarity).} A lightweight autoencoder maps drug embeddings to a latent $z$ and decodes token logits over the SMILES vocabulary. We train with token-level cross-entropy (ignoring \textsc{pad} tokens):
\[
L_{\text{Recon}}
= - \frac{1}{T_{\text{eff}}}\sum_{t \in \text{non-PAD}}
\log p(s_t \mid z),
\]
where $T_{\text{eff}}$ counts non-\textsc{pad} positions. Unless otherwise stated, we use
$\alpha_{\text{cls}}{=}0.4$,
$\alpha_{\text{con}}{=}0.2$,
$\alpha_{\text{conf}}{=}0.2$,
$\alpha_{\text{recon}}{=}0.2$, as this configuration yielded the best validation performance among the tested weightings.

\section{Ablation Studies on Embedding Effectiveness} \label{ablation}
\subsection{Ablation Studies for DTI} \label{ablation_dti}

We conducted extensive ablation experiments on widely used DTI benchmarks (BIOSNAP, BindingDB, and DAVIS) to identify the most effective drug and protein embeddings for accurate interaction prediction. Below, we present our findings and the rationale for the selected configurations.

\subsubsection{Drug Embeddings}

We evaluated multiple drug embedding strategies to determine the most effective representation for our model:
\begin{itemize}
    \item \textbf{Mollm}: A transformer-based model that extracts molecular features through self-attention mechanisms \citep{vaswani2017attention, radford2021learning}.
    \item \textbf{GIN with Barlow Twins Loss}: A self-supervised method using Graph Isomorphism Networks to learn robust molecular representations \citep{zbontar2021barlow}.
    \item \textbf{Cooperative Protocol (COOP)}: An embedding approach integrating cooperative strategies to improve drug representations \citep{finkelshtein2023cooperative}.
    \item \textbf{Molecular FingerPrints (MFPS)}: Provides robust and detailed encodings of molecular structures by converting molecules into fixed-length binary vectors representing the presence or absence of particular substructures \citep{mfps}.
    \item \textbf{Graph Convolutional Network (GCN)}: A neural network model that effectively captures the structural features of drugs by transforming molecular graphs into high-dimensional embeddings suitable for interaction prediction \citep{kipf2016semi}.
\end{itemize}

For the initial ablation study, we used ESM-2 for protein embeddings and tested different drug embeddings. The results are presented in Table \ref{table:drug_embeddings}.

\begin{table*}[ht]
\centering
\caption{Performance of Tensor-DTI on different datasets with various drug embeddings. $GIN_L$ differs from GIN by using a larger training dataset and a more complex architecture. All values correspond to AUPR.}
\vspace{0.15in}
\begin{adjustbox}{max width=\textwidth}
\begin{tabular}{lccccccc}
\hline
Dataset/d. emb. & GCN & $GIN_L$ & MFPS & GIN & COOP\\
\hline
BIOSNAP & \underline{0.879} & 0.832 & \textbf{0.881} & 0.837 & 0.837 \\
unseen T & \underline{0.708} & 0.646 & \textbf{0.720} & 0.649 & 0.638 \\
unseen D & \textbf{0.872} & 0.827 & \underline{0.851} & 0.832 & 0.832 \\
BindingDB & \underline{0.664} & 0.583 & \textbf{0.679} & 0.591 & 0.581 \\
DAVIS & \textbf{0.532} & 0.331 & \underline{0.527} & 0.334 & 0.338\\
\hline
\end{tabular}
\end{adjustbox}
\label{table:drug_embeddings}
\end{table*}

GCN performed particularly well on unseen drugs, while MFPS achieved the highest overall AUPR scores across benchmarks. These findings confirm their robustness in different settings, leading to their selection for further analysis.

\subsubsection{Protein Embeddings}

After evaluating drug embeddings, we assessed the impact of protein embeddings, comparing SaProt and ESM-2 to determine their effect on model performance.

\begin{itemize}
    \item \textbf{SaProt Embeddings}: Derived from a transformer-based model specifically designed for protein sequences, offering high-quality embeddings \citep{saprot}.
    \item \textbf{ESM-2 Embeddings}: Generated by a state-of-the-art transformer model trained on a large corpus of protein sequences, known for its robust performance \citep{esm2}.
\end{itemize}

The results of this ablation study are shown in Table \ref{table:protein_embeddings}.

\begin{table*}[ht]
\centering
\caption{Performance of Tensor-DTI on different datasets with various protein embeddings.}
\vspace{0.15in}
\begin{adjustbox}{max width=\textwidth}
\begin{tabular}{lcccccccc}
\hline
Dataset/d. emb. & GCN (ESM-2) & MFPS (ESM-2) & GCN (SaProt) & MFPS (SaProt)\\
\hline
BIOSNAP & 0.879 & 0.881 & \textbf{0.897} & \underline{0.894}\\
unseen T & 0.708 & 0.720 & \underline{0.836} & \textbf{0.838}\\
unseen D & \underline{0.872} & 0.851 & \textbf{0.879} & 0.849\\
BindingDB & 0.664 & 0.679 & \underline{0.685} & \textbf{0.689}\\
DAVIS & 0.532 & 0.527 & \textbf{0.555} & \underline{0.552}\\
\hline
MEAN & 0.731 & 0.732 & \textbf{0.770} & \underline{0.764} \\
\hline
\end{tabular}
\end{adjustbox}
\label{table:protein_embeddings}
\end{table*}

The results indicated that SaProt embeddings consistently outperformed ESM-2 embeddings across multiple datasets, leading us to select SaProt for protein embeddings in our final model.
We also selected GCN as the technique for further analysis.

\subsubsection{Impact of Training the GCN}

To assess the impact of large-scale training, we examined whether pretraining the GCN on a larger dataset (PCBA\_1328) improves performance. As shown in Table \ref{table:gcn_trained}, the pretrained GCN consistently outperforms its untrained counterpart, achieving higher AUPR scores across all benchmarks.

\begin{table*}[h!]
\centering
\caption{Performance of Tensor-DTI with Trained GCN embeddings and ConPlex.}
\vspace{0.15in}
\begin{adjustbox}{max width=\textwidth}
\begin{tabular}{lccc}
\hline
Dataset/d. emb. & GCN & \textbf{Trained GCN} & ConPlex(MFPS) \\
\hline
BIOSNAP & \underline{0.900 ± 0.002} & \textbf{0.903 ± 0.003} & 0.897 ± 0.001 \\
unseen T & 0.834 ± 0.004 & \underline{0.839 ± 0.003} & \textbf{0.842 ± 0.006} \\
unseen D & \underline{0.880 ± 0.004} & \textbf{0.888 ± 0.002} & 0.874 ± 0.002 \\
BindingDB & \underline{0.686 ± 0.003} & \textbf{0.699 ± 0.002} & 0.628 ± 0.012 \\
DAVIS & \underline{0.544 ± 0.015} & \textbf{0.547 ± 0.006} & 0.458 ± 0.016 \\
\hline
\end{tabular}
\end{adjustbox}
\label{table:gcn_trained}
\end{table*}

These findings demonstrate that both GCN and trained GCN embeddings significantly outperform ConPlex in most benchmarks, reinforcing the robustness of our proposed Tensor-DTI model.

\subsection{Ablation Studies for DTA} \label{ablation_dta}

In addition to our comprehensive analysis for DTI, we conducted ablation studies for DTA prediction in the TDC-DG benchmark to identify the optimal embeddings for both drugs and proteins.

\subsubsection{Protein Embeddings}

We first compared the performance of different protein embeddings, specifically SaProt and ESM-2 embeddings:

\begin{itemize}
    \item \textbf{SaProt Embeddings} \citep{saprot}.
    \item \textbf{ESM-2 Embeddings} \citep{esm2}.
\end{itemize}

The results of this ablation study are shown in Table \ref{table:protein_embeddings_dta}.

\begin{table*}[ht]
\centering
\caption{Performance of Tensor-DTI in terms of PCC on TDC-DG with two different protein embeddings.}
\vspace{0.15in}
\begin{adjustbox}{max width=\textwidth}
\begin{tabular}{lcccc}
\hline
t. emb./d. emb. & GCN \\
\hline
ESM-2 & \textbf{0.550}\\
SaProt-650M & 0.530\\
\hline
\end{tabular}
\end{adjustbox}
\label{table:protein_embeddings_dta}
\end{table*}

Based on these results, ESM-2 embeddings were selected due to their superior performance.

\subsubsection{Drug Embeddings}

We then evaluated two primary drug embedding methods for our DTA model:
\begin{itemize}
    \item \textbf{Molecular FingerPrints (MFPS)} \citep{mfps}.
    \item \textbf{Graph Convolutional Network (GCN)} \citep{kipf2016semi}.
\end{itemize}

For this ablation study, we used ESM-2 for protein embeddings and evaluated different drug embeddings. The results are presented in Table \ref{table:drug_embeddings_dta}.

\begin{table*}[ht]
\centering
\caption{Performance of Tensor-DTI and ConPlex on different datasets with various drug embeddings (PCC).}
\vspace{0.15in}
\begin{adjustbox}{max width=\textwidth}
\begin{tabular}{lcccccc}
\hline
-/d. emb. & GCN & MFPS & Trained GCN & ConPlex (MFPS) \\
\hline
PCC & \underline{0.546 ± 0.02} & \textbf{0.580 ± 0.004} & 0.539 ± 0.001 & 0.538 ± 0.008 \\
\hline
\end{tabular}
\end{adjustbox}
\label{table:drug_embeddings_dta}
\end{table*}

Among the evaluated methods, MFPS achieved the highest PCC score (0.580), demonstrating superior performance compared to other embedding strategies. Given its consistently strong results across benchmarks, we selected MFPS as the preferred drug embedding for the final model.

Additionally, Table \ref{table:drug_embeddings_dta} confirms that in this case, pretraining the GCN did not provide any advantage for the DTA study. This may be attributed to the activity information contained in the PCBA\_1328 dataset.

\subsection{Ablation Studies for DTA - Leak Proof Benchmark} \label{ablation_dta_lp}

To further assess the impact of embedding choices on DTA prediction under strict data leakage constraints, we evaluated Tensor-DTI on the LP-PDBBind benchmark. The results highlight significant differences in performance based on the selected embeddings.

The best-performing configuration combined SaProt protein embeddings with trained GCN drug embeddings, achieving a PCC of \(0.565\) and an RMSE of \(1.62\). In contrast, using ESM-2 protein embeddings with MFPS drug embeddings led to a lower PCC of \(0.450\) and a higher RMSE of \(1.79\). Overall, the best-performing configuration combined SaProt protein embeddings with trained GCN drug embeddings, achieving the highest PCC and lowest RMSE. These findings highlight the importance of structural protein embeddings (SaProt) and graph-based drug representations (trained GCN) in improving model generalization under strict data leakage constraints. This underscores the critical role of domain-specific, structure-informed embeddings in achieving robust and accurate affinity predictions in real-world applications.

\section{Databases}\label{databases}

Data collection, processing and splitting are pivotal in drug-target interaction predictions. We outline all datasets used in this work with detailed descriptions on the train-validation-test splittings performed over them.

\textbf{BIOSNAP.} This drug-target interaction network provides information on the genes (i.e., proteins encoded by genes) targeted by drugs available on the U.S. market. Drug targets are molecules essential for the transport, delivery, or activation of a drug. BIOSNAP information is widely utilized in computational drug target discovery, drug design, docking or screening, metabolism prediction, interaction prediction, and general pharmaceutical research. Drug entries span small molecules, biologics, and nutraceutical compounds. On average, drugs have 5-10 unique target proteins. The dataset lists all known targets with physiological or pharmaceutical effects, not just a single primary target, and fully accounts for the fact that many targets are protein complexes composed of multiple subunits or combinations of proteins. 

\textit{Preprocessing and splitting.} We use ChGMiner from BIOSNAP, which contains only positive drug-target interactions. Following the approach described in \citep{ConPlex}, we create negative DTIs by randomly sampling an equal number of protein-drug pairs, under the assumption that a random pair is unlikely to interact positively.

\textbf{DAVIS.} The DAVIS dataset is a comprehensive resource profiling interactions between 72 kinase inhibitors and 442 kinases, covering over 80\% of the human catalytic protein kinome. It provides detailed binding affinity data (\(K_d\)) for each interaction and calculates selectivity scores to evaluate inhibitor specificity. The dataset distinguishes between type I inhibitors, which target active kinase conformations, and type II inhibitors, which bind inactive states, showing that type II inhibitors are generally more selective, though exceptions exist. It highlights group-selective inhibitors, off-target profiles, and structural features contributing to selectivity, making it invaluable for drug discovery, kinase biology, and computational modeling.

\textbf{BindingDB.} BindingDB is a publicly accessible database that provides experimentally determined protein-ligand binding affinities, focusing on drug-target interactions. It currently contains over 20\,000 binding measurements for approximately 11\,000 small molecule ligands and 110 protein targets, including isoforms and mutants. BindingDB integrates data from enzyme inhibition studies and isothermal titration calorimetry, extracted from scientific literature and directly deposited by experimentalists. The database is designed to support diverse applications, such as computational drug design, ligand discovery, and structure-activity relationship analysis. Its web interface offers powerful tools for querying by chemical structure, substructure, protein sequence, or affinity ranges, and supports virtual screening using uploaded compound databases. By linking data to the Protein Data Bank (PDB) and PubMed, BindingDB facilitates the integration of binding, structural, and sequence data, making it a valuable resource for researchers in pharmaceutical sciences and computational biology.

\textit{Preprocessing and splitting for DAVIS and BindingDB datasets.} Following the approach described in \citep{ConPlex}, we treat pairs with \(K_d\) $<$ 30 as positive DTIs and those with larger \(K_d\) values as negative DTIs. The dataset is split into 70\% for training, 10\% for validation, and 20\% for testing. Training data are subsampled to have an equal number of positive and negative interactions, ensuring a balanced training set, while validation and test data retain the original ratio of interactions. This preprocessing strategy ensures consistency across datasets and facilitates robust model evaluation. Compared to DAVIS, which represents a low-data learning setting with 2\,086 training interactions, BindingDB provides a broader learning scenario with 12\,668 training interactions, offering greater diversity in drug-target interaction pairs. Both datasets complement each other, enabling evaluation of model performance across varying levels of data availability.

\textbf{DUD-E (Kinase Subset).} DUD-E provides a curated set of protein targets and their known binding partners, along with decoy molecules designed to resemble the physicochemical properties of true binders but are experimentally confirmed not to bind. From this database, we focus specifically on the kinase family, which consists of 26 protein targets. For each target, the dataset includes an average of 224 true binding partners and 50 decoys per binding partner, enabling robust evaluation of model performance in distinguishing between true and decoy interactions. 

\textit{Preprocessing and splitting.} We derive train-test splits by partitioning the kinase targets such that no target appears in both the training and test sets. Specifically, we hold out 50\% of kinase targets for testing and use the remaining targets for training, ensuring representative coverage of the kinase family in both splits. This setup evaluates the ability of models to generalize to unseen kinase targets and to distinguish true binding interactions from decoy molecules.

\textbf{PLINDER.} PLINDER \citep{plinder} is a comprehensive protein-ligand interaction dataset designed to address critical challenges in computational drug design and protein engineering. It includes 449\,383 systems, each extensively annotated with over 500 attributes, including protein, ligand, pocket, and interaction-level similarity metrics. PLINDER uniquely links holo systems to apo and predicted structures, enabling realistic evaluation scenarios such as docking, ligand generation, and co-folding. By employing advanced splitting algorithms, PLINDER minimizes information leakage, enhancing the evaluation of machine learning models’ generalization capabilities. Its rich annotations, task-specific test sets, and robust evaluation frameworks make PLINDER a valuable resource for advancing predictive modeling in protein-ligand interactions.

\textit{Preprocessing and splitting.} PLINDER preprocesses and splits its extensive protein-ligand interaction (PLI) dataset, comprising 449\,383 systems from 162\,978 PDB entries, into training, validation, and test sets with rigorous quality control and annotation. A total of 113\,498 high-quality systems meeting stringent criteria such as resolution $\leq$ 3.5 Å and R-factor $\leq$ 0.4 are annotated with over 500 features, including protein, pocket, ligand, and interaction-level details. Train-test splits are generated using graph-based algorithms that classify systems based on similarity metrics like protein sequence identity, pocket overlap, and ligand fingerprints, ensuring minimal leakage and maximum diversity. The PLINDER-PL50 split includes 57\,602 training, 3\,453 validation, and 3\,729 test systems, achieving 0\% leakage for PLI similarity $\geq$ 50\%. Another configuration, PLINDER-ECOD, defines splits using evolutionary domains and comprises 77\,411 training, 10\,169 validation, and 12\,174 test systems, all containing 100\% high-quality systems to support robust and realistic benchmarking of computational models.

\textbf{SMPBind-I.} The following dataset is a curated mix of several databases. These databases include ChEMBL, PubChem, ChEBI (Chemical Entities of Biological Interest), STITCH, OpenTargets, DGIdb (Drug Gene Interaction Database), Pharos, TTD (Therapeutic Targets Database), HMDB (Human Metabolome Database), T3DB (Toxin and Toxin-Target Database), BindingDB and DTC (Drug Target Commons). 

\textit{Preprocessing and splitting.} From these databases we extract the pairs of protein molecules that have been experimentally validated at least once. Afterwards, we performed an extensive de-duplication procedure. Racemic mixtures are separated into their chiral parts, hydrogen atoms are removed, metal atoms are disconnected from the molecule, the molecule is normalized and reionized. After this point, in the case that the molecule has several fragments, the biggest one is assumed to be the bioactive one, so it is selected. Then the molecule is neutralized and canonicalized, to avoid the presence of tautomerism overlap within the database. Lastly, InChIKeys are computed from the resulting molecules and used for de-duplication. The resulting database contains more than 400\,000 different Murcko scaffolds, and more than 35\,000 unique proteins, divided in over 7\,000 different families and 1\,000 superfamilies.

\textbf{Propedia.} Propedia v2.3 \citep{propedia} is a peptide-protein interaction database. The last updated version builds on the foundational Propedia database by incorporating over 49\,300 peptide-protein complexes—a 150\% increase from its initial release—and introducing graph-based structural signatures to represent peptide structures numerically. These signatures, calculated using the aCSM-ALL algorithm, enhance the ability to cluster and analyze peptides based on sequence similarities, structural interfaces, and binding sites. Propedia v2.3 supports machine learning applications, offering a CSV dataset of feature vectors suitable for tasks like peptide classification and therapeutic discovery. The database facilitates in-depth exploration of peptide-protein recognition mechanisms, a critical aspect of drug development and biotechnology. 

\textit{Preprocessing and splitting.} We preprocess the Propedia dataset by clustering protein and peptide embeddings into distinct groups using K-Means. Protein and peptide embeddings are numerically represented. Clustering ensures that similar protein and peptide structures are grouped together, facilitating representative splitting across training, validation, and test sets.
The dataset is split into 80\% for training, 10\% for validation, and 10\% for testing, ensuring that no peptide-protein pairs from the same cluster appear across different splits. Negative pairs are generated by randomly sampling an equal number of peptide-protein pairs within each split, resulting in a 1:1 balance of positive and negative interactions in all sets. The original dataset contains only positive pairs, and this augmentation ensures balanced training and evaluation.

\textbf{CoPRA dataset.} The CoPRA dataset \citep{CoPRA} is designed for protein-RNA binding affinity prediction and consists of two subsets: PRI30k (training), and PRA310 (test). This dataset is directly provided by CoPRA and includes 30\,000 non-redundant protein-RNA complexes from BioLiP2 (PRI30k) and 310 high-quality complexes curated from PDBBind, PRBABv2, and ProNAB (PRA310).

\textit{Preprocessing and splitting.} Positive interactions are extracted from experimental annotations, while negative pairs are generated by random pairing within each subset to ensure a 1:1 positive-negative ratio. Clusters are created using CD-HIT at 70\% sequence identity to prevent data leakage, with distinct splits for training, validation, and testing. This setup ensures diverse, high-quality data for robust model evaluation \citep{CoPRA}.

\newpage
\section{Overview of Dataset Sizes Across Benchmarks} \label{dataset_volumes}

\begin{table*}[ht!]
\centering
\caption{Dataset sizes across splits. Cells show Positive/Negative pairs. DUD-E has no validation split. (p) stands for the pocket split of the PLINDER dataset.}
\label{tab:dataset_volumes}
\setlength{\tabcolsep}{4pt}
\renewcommand{\arraystretch}{1.08}
\begin{adjustbox}{max width=\textwidth}
\begin{tabular}{llccc}
\toprule
\textbf{Target–Ligand} & \textbf{Benchmark} & \textbf{Train (P/N)} & \textbf{Validation (P/N)} & \textbf{Test (P/N)} \\
\midrule
\multirow{7}{*}{Protein-Drug}
  & BIOSNAP         & 9\,490/9\,306 & 1\,372/1\,327 & 2\,718/2\,656 \\
  & BindingDB       & 5\,842/5\,702 &   859/5\,134   & 1\,752/10\,307 \\
  & DAVIS           &   883/909       &   132/2\,474   &   252/4\,987 \\
  & DUD-E (Kin.)  & 4\,112/150\,027 & \textemdash{} & 5\,027/201\,599 \\
  & PLINDER         & 400\,351/400\,351 & 31\,612/31\,612 & 27\,775/27\,775 \\
  & PLINDER (p)& 134\,909/134\,909 & 6\,789/6\,789 & 5\,645/5\,645 \\
  & SMPBind-I         & 10\,340\,470/10\,349\,292 & 1\,292\,375/1\,293\,821 & 1\,292\,591/1\,293\,639 \\
\midrule
Protein-Peptide& Propedia & 40\,370/40\,393 & 4\,418/4\,419 & 4\,415/4\,412 \\
Protein-RNA& CoPRA    & 15\,626/15\,626 &   820/820       &   200/200 \\
\bottomrule
\end{tabular}
\end{adjustbox}
\end{table*}

\begin{table*}[ht!]
\centering
\caption{Dataset sizes across splits for DTA tasks. Cells show positive pairs only.}
\label{tab:dta_volumes}
\setlength{\tabcolsep}{4pt}
\renewcommand{\arraystretch}{1.08}
\begin{tabular}{llccc}
\toprule
\textbf{Target-Ligand} & \textbf{Benchmark} & \textbf{Train} & \textbf{Validation} & \textbf{Test} \\
\midrule
\multirow{5}{*}{Protein-Drug}
  & TDC-DG                 & 146\,744 & 36\,686 & 49\,028 \\
  & LP-PDBBIND             & 5\,691   & 1\,317  & 3\,103 \\
  & LP-PDBBIND (\(\Delta G\)) & 5\,691   & 1\,317  & 3\,103 \\
  & PDBBind-Opt            & 13\,185  & 1\,465  & 1\,628 \\
  & PDBBind-Opt+LP         & 7\,051   & 1\,846  & 4\,193 \\
\midrule
Protein-Peptide & PDBBind-Opt & 1\,896 & 210 & 240 \\
RNA-Drug        & PDBBIND     & 96 & 13 & 11 \\
Protein-RNA     & CoPRA       & 165 & 21 & 14 \\
\bottomrule
\end{tabular}
\end{table*}

\newpage
\section{Low-Leakage Datasets Results}\label{low_leak_results}

To assess Tensor-DTI’s robustness in minimized leakage scenarios, we compare its performance against a one-hot encoding baseline across multiple datasets, ensuring consistency in hyperparameter settings.

\begin{table*}[h]
    \centering
    \caption{Performance comparison of DTI (classification) and DTA (regression) models on minimized leakage datasets. AUPR is used for classification benchmarks, while PCC and RMSE evaluate affinity prediction tasks.}
    \vspace{0.15in}
    \renewcommand{\arraystretch}{1.2}
    \begin{adjustbox}{max width=\textwidth}
    \begin{tabular}{lcccc}
        \toprule
        \textbf{Benchmark} & \textbf{Model} & \textbf{AUPR} & \textbf{PCC} & \textbf{RMSE} \\
        \midrule
        PLINDER (no pocket) & Tensor-DTI & $0.785 \pm 0.002$ & - & - \\
        PLINDER (no pocket | pocket data) & Tensor-DTI & $0.739 \pm 0.005$ & - & - \\
        PLINDER (pocket | pocket data) & Tensor-DTI & $0.754 \pm 0.005$ & - & - \\
        \midrule
        LP-PDBBind & &  &  &  \\
         & Tensor-DTI & - & $0.565 \pm 0.004$ & $1.620 \pm 0.024$ \\
         & DeepDTA & - & $0.512 \pm 0.020$ & $2.290 \pm 0.040$ \\
         & AutoDock Vina & - & $0.450 \pm 0.000$ & $2.560 \pm 0.000$ \\
         & One-Hot & - & $0.428 \pm 0.016$ & $2.287 \pm 0.032$ \\
       LP-PDBBind ($\Delta G$ prediction) & &  & &  \\
        & Tensor-DTI & - & $0.528 \pm 0.013$ & $2.122 \pm 0.032$ \\
        & One-Hot & - & $0.428 \pm 0.016$ & $2.287 \pm 0.032$ \\
        \midrule
        PDBBind-Opt Peptide-Protein &  &  &  & \\
         & Tensor-DTI &-  & $0.679 \pm 0.014$ & $1.175 \pm 0.020$ \\
        & One-Hot & - & $0.568 \pm 0.025$ & $1.846 \pm 0.099$ \\
        PDBBind-Opt Small Molecule-Protein &  &  &  & \\
         & Tensor-DTI & - & $0.750 \pm 0.005$ & $1.335 \pm 0.011$ \\
        & One-Hot & - & $0.728 \pm 0.007$ & $1.320 \pm 0.012$ \\
        \multirow{3}{*}{\shortstack{PDBBind-Opt Small Molecule-Protein\\(Leak proof split)}} &  &  &  & \\
         & Tensor-DTI & - & $0.493 \pm 0.005$ & $1.545 \pm 0.006$ \\
        & One-Hot & - & $0.385 \pm 0.014$ & $1.752 \pm 0.033$ \\
        \bottomrule
    \end{tabular}
    \end{adjustbox}
    \label{tab:low_leakage_performance}
\end{table*}

\newpage
\section{Expanded Results on Biomolecular Interaction Predictions}\label{biomolecular_pred}

We compare Tensor-DTI against a one-hot encoding baseline under the same hyperparameter settings to evaluate its performance across various biomolecular interaction tasks.

\begin{table}[h!]
\centering
\caption{Performance of Tensor-DTI on the Propedia peptide interaction database.}
\vspace{0.15in}
\label{tab:tensor_dti_propedia}
\begin{tabular}{@{}lc@{}}
\toprule
\textbf{Model}     & \textbf{AUPR}  \\ \midrule
Tensor-DTI           & $0.953 \pm 0.001$          \\
One-hot encoding           & $0.884 \pm 0.003$          \\
\bottomrule
\end{tabular}
\end{table}

\begin{table}[h!]
\centering
\caption{Performance of Tensor-DTI on CoPRA}
\vspace{0.15in}
\label{tab:tensor_dti_copra}
\begin{tabular}{@{}lc@{}}
\toprule
\textbf{Model}  & \textbf{AUPR} \\ \midrule
Tensor-DTI       & $0.916 \pm 0.008$       \\ 
One-hot encoding           & $0.795 \pm 0.009$          \\\bottomrule
\end{tabular}
\end{table}

\begin{table}[h!]
\centering
\setlength{\tabcolsep}{6pt} 
\renewcommand{\arraystretch}{1.3} 
\caption{Performance of Tensor-DTI on the CoPRA (PRA310). The top table corresponds to \(K_d\) (binding constant) prediction, while the bottom table corresponds to \(\Delta G\) (free energy) prediction.}
\vspace{0.15in}
\label{tab:protein_rna}
\begin{tabular}{lcc}
\multicolumn{3}{c}{\textbf{\(K_d\) Prediction}} \\ 
\midrule
\textbf{Model} & \textbf{PCC} & \textbf{RMSE} \\ 
\midrule
Tensor-DTI & $0.631 \pm 0.111$ & $1.443 \pm 0.232$ \\ 
One-hot encoding & $0.468 \pm 0.189$ & $1.399 \pm 0.232$ \\ 
\bottomrule
\end{tabular}

\vspace{0.5cm} 

\begin{tabular}{lcc}
\multicolumn{3}{c}{\textbf{\(\Delta G\) Prediction}} \\ 
\midrule
\textbf{Model} & \textbf{PCC} & \textbf{RMSE} \\ 
\midrule
Tensor-DTI & $0.621 \pm 0.052$ & $1.910 \pm 0.212$ \\ 
One-hot encoding & $0.453 \pm 0.213$ & $1.896 \pm 0.430$ \\ 
\bottomrule
\end{tabular}
\end{table}

\begin{table}[h!]
\centering
\setlength{\tabcolsep}{5pt} 
\caption{Performance of Tensor-DTI on the PDBBind interaction database, selecting from it the Drug-RNA interactions.}
\vspace{0.15in}
\label{tab:drug_rna}
\begin{tabular}{lcc}
\toprule
\textbf{Model} & \textbf{PCC} & \textbf{RMSE} \\ 
\midrule
Tensor-DTI & $0.792 \pm 0.015$ & $1.684 \pm 0.038$ \\ 
One-hot encoding & $0.633 \pm 0.018$ & $1.738 \pm 0.036$ \\ 
\bottomrule
\end{tabular}
\end{table}

\end{document}